\documentclass[10pt,twocolumn,letterpaper]{article}

\usepackage{cvpr}              % To produce the CAMERA-READY version
\definecolor{cvprblue}{rgb}{0.21,0.49,0.74}
\usepackage[pagebackref,breaklinks,colorlinks,allcolors=cvprblue]{hyperref}

\usepackage{hyperref}
\usepackage{url}

\usepackage{listings} % For lstlisting
\usepackage{color} % For defining colors
\usepackage{xspace} % For defining colors
\usepackage{fancyvrb}
\usepackage{fvextra} % Extended features for fancyvrb
\usepackage[most]{tcolorbox}

\lstdefinestyle{mystyle}{
    columns=fullflexible,
    backgroundcolor=\color{backcolour},   
    commentstyle=\color{codegreen},
    numberstyle=\tiny\color{codegray},
    stringstyle=\color{codepurple},
    basicstyle=\footnotesize,
    breakatwhitespace=false,         
    breaklines=true,  
    breakindent=0pt,
    captionpos=b,                    
    keepspaces=true,  
    numbersep=0pt,                  
    showspaces=false,                
    showstringspaces=false,
    showtabs=true,                  
    tabsize=10,
    escapeinside={@}{@}, % Define escape sequence for LaTeX inside listings
    moredelim=**[is][\highlight]{`}{`}, % Highlight words with `
}
\usepackage[T1]{fontenc}
\usepackage[most]{tcolorbox}

\usepackage{tocloft} % for TOC management
\usepackage{etoolbox}

\usepackage[utf8]{inputenc} % allow utf-8 input
\usepackage[T1]{fontenc}    % use 8-bit T1 fonts
\usepackage{hyperref}       % hyperlinks
\usepackage{url}            % simple URL typesetting
\usepackage{booktabs}       % professional-quality tables
\usepackage{amsfonts}       % blackboard math symbols
\usepackage{nicefrac}       % compact symbols for 1/2, etc.
\usepackage{microtype}      % microtypography
\usepackage{xcolor}         % colors

\usepackage{graphicx}
\usepackage{multirow}
\usepackage{enumitem}

\usepackage{color} % For defining colors

\RequirePackage{xspace}
\makeatletter
\DeclareRobustCommand\onedot{\futurelet\@let@token\@onedot}
\def\@onedot{\ifx\@let@token.\else.\null\fi\xspace}
\def\eg{\emph{e.g}\onedot} 

\def\ie{\emph{i.e}\onedot}

\def\wrt{w.r.t\onedot}

\definecolor{codegreen}{rgb}{0,0.6,0}
\definecolor{codegray}{rgb}{0.5,0.5,0.5}
\definecolor{codepurple}{rgb}{0.58,0,0.82}
\definecolor{backcolour}{rgb}{0.95,0.95,0.92}
\newcommand{\highlight}[1]{\color{blue}{#1}}
\lstdefinestyle{mystyle}{
    columns=fullflexible,
    backgroundcolor=\color{backcolour},   
    commentstyle=\color{codegreen},
    numberstyle=\tiny\color{codegray},
    stringstyle=\color{codepurple},
    basicstyle=\footnotesize,
    breakatwhitespace=false,         
    breaklines=true,  
    breakindent=0pt,
    captionpos=b,                    
    keepspaces=true,  
    numbersep=0pt,                  
    showspaces=false,                
    showstringspaces=false,
    showtabs=true,                  
    tabsize=10,
    escapeinside={@}{@}, % Define escape sequence for LaTeX inside listings
    moredelim=**[is][\highlight]{`}{`}, % Highlight words with `
}
\usepackage{colortbl}

\newcommand{\AL}[1]{{\color{magenta}{[Andrew: #1]}}}

\newcommand{\DA}[1]{{\color{red}{[David: #1]}}}
\newcommand{\RRM}[1]{{\color{green}{[Rafid: #1]}}}
\newcommand{\new}[1]{{#1}}
\newcommand{\iclr}[1]{{#1}}
\newcommand{\neurips}[1]{{#1}}%{{\color{blue}{#1}}}
\newcommand{\cvpr}[1]{{#1}}
\usepackage{amsmath,amsfonts,bm}

\newcommand{\figleft}{{\em (Left)}}

\def\eqref#1{equation~\ref{#1}}
\def\1{\bm{1}}

\def\rh{{\textnormal{h}}}
\def\ri{{\textnormal{i}}}

\def\ro{{\textnormal{o}}}

\def\rq{{\textnormal{q}}}
\def\rr{{\textnormal{r}}}

\def\rt{{\textnormal{t}}}

\def\rw{{\textnormal{w}}}
\def\rx{{\textnormal{x}}}

\DeclareMathAlphabet{\mathsfit}{\encodingdefault}{\sfdefault}{m}{sl}
\SetMathAlphabet{\mathsfit}{bold}{\encodingdefault}{\sfdefault}{bx}{n}

\newcommand{\R}{\mathbb{R}}

\def\rN{{\textnormal{N}}}

\def\paperID{16911} % *** Enter the Paper ID here
\def\confName{CVPR}
\def\confYear{2025}

\title{Can Large Vision-Language Models Correct Semantic Grounding Errors By Themselves?}

\author{
Yuan-Hong Liao\textsuperscript{1}, Rafid Mahmood\textsuperscript{2,3}, Sanja Fidler\textsuperscript{1,2}, David Acuna\textsuperscript{2} \\
\textsuperscript{1}University of Toronto, Vector Institute \textsuperscript{2}NVIDIA \textsuperscript{3}University of Ottawa \\
{\tt\small answer@cs.toronto.edu, mahmood@telfer.uottawa.ca, \{sfidler, dacunamarrer\}@nvidia}
}

\begin{document}
\maketitle

\begin{abstract}
Enhancing semantic grounding abilities in Vision-Language Models (VLMs) often involves collecting domain-specific training data, refining the network architectures, or modifying the training recipes. In this work, we venture into an orthogonal direction and explore 
% to improve 
semantic grounding in VLMs through \emph{self-correction}, without requiring in-domain data, fine-tuning, or modifications to the network architectures.
Despite the concerns raised in the self-correction of LLMs, we find that if prompted and framed properly, VLMs \emph{can} correct their own semantic grounding mistakes even without the access to the oracle feedback. 
%
% Finally, we adopt the identified self-correction framework in an iterative setting which \emph{consistently} improve across all models investigated. Overall, through iterative self-correction, VLMs improve up to 8.4 accuracy points.
We also show an identified self-correction framework in an iterative setting which \emph{consistently} improves performance across all models investigated. 
% Overall, through iterative self-correction, VLMs improve up to 8.4 accuracy points.
Overall, iterative self-correction consistently improves VLM performance by up to 8.4 accuracy points across all models investigated; yet, after several rounds of feedback, strong models like GPT-4V and GPT-4o still exhibit significant error rates, indicating promising directions for further research.

\end{abstract}

\section{Introduction}
\vspace{-1mm}

The evolution of Large Language Models (LLMs) to encompass multimodal inputs has given rise to an emerging paradigm of general-purpose models that can solve multimodal understanding problems via user-prompt interaction~\citep{touvron2023llama,reid2024gemini,geminiteam2023gemini,gpt4v_report,mckinzie2024mm1}. %\DA{This is not the best citation to put here, cite GPT4,Gemini,LLama,apple's mm1} \AL{added}
Vision-Language Models (VLMs) are a growing family of multimodal models that simultaneously understand both visual and language cues. These models have demonstrated strong zero-shot performance on tasks including image classification~\citep{deng2009imagenet}, captioning~\citep{young2014image}, visual question answering~\citep{VQA,goyal2017making}, %\DA{wrong reference?} 
reasoning
% from image data
~\citep{refcoco,clip_bag_of_words} 
%\DA{this references are ok but probably for common sense  not the most appropiate, cite COCO, referCOCO, referCOCO}
 % motivating 
 and
 % application-specific VLMs 
 % in
 % autonomous vehicles and 
 robotics~\citep{cui2024survey, pivot}.

\begin{figure*}[t]
\includegraphics[width=\textwidth]{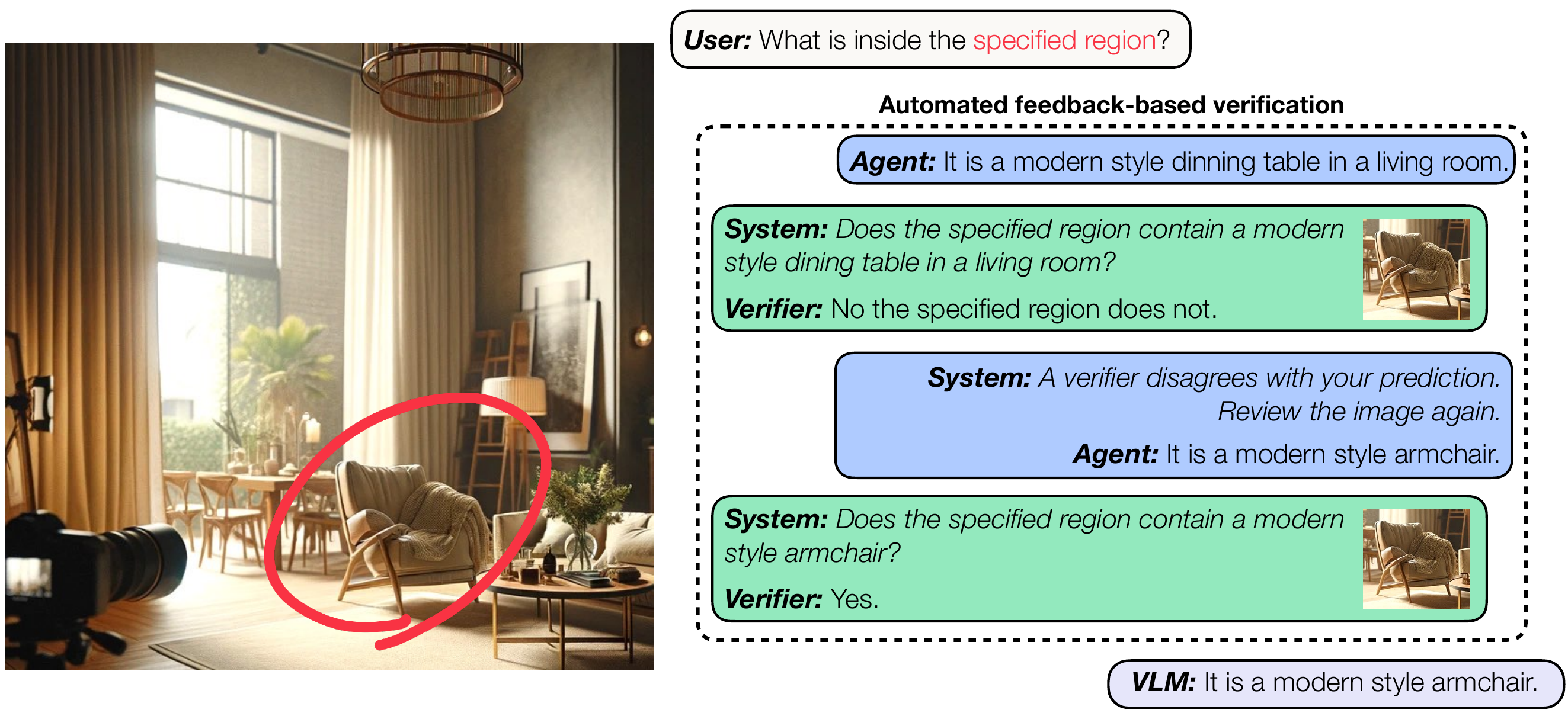}
%\caption{\textbf{Enhancing semantic grounding in VLMs with feedback.} 
\caption{\textbf{Enhancing semantic grounding in VLMs \iclr{through self-correction}.} 
\iclr{We explore to improve semantic grounding in VLMs through self-correction, without the needs of in-domain data, fine-tuning, or architectural changes. For self-correction, we adopt the setup involving explicit feedback generation.}
%We explore if VLMs can enhance their grounding performance using feedback alone, without needing in-domain data, fine-tuning, or architectural changes. 
When provided with an image and a specified region, a VLM identifies the semantic properties of the image region.
An automated feedback-based verification mechanism facilitates an interaction between the VLM and a `Verifier' to improve the VLM's initial understanding.
}
\label{fig:teaser}
\vspace{-3mm}
\end{figure*}

\iclr{
Despite VLMs' strong visual-language understanding abilities, fine-grained visual grounding remains challenging. Specifically, VLMs struggle to comprehend region-specific information within complex scenes, such as when prompted to describe specific objects in a crowded image~\citep{chen2023shikra, som_prompt, you2023ferret} (see Fig.~\ref{fig:teaser}). Prior works address this limitation using additional in-domain data~\citep{guo2024regiongpt, lin2023vila, li2023otterhd}, finetuning, or architectural changes~\citep{li2024monkey, llava_1.6}. However, these approaches require significant computational resources~\citep{vip_llava, you2023ferret}. %Thus, enhancing VLMs for fine-grained visual grounding without substantial computational overhead remains an open challenge. \DA{consider removing the previous sentence in this new framing}
}

On the other hand, adjacent LLM literature has demonstrated that LLMs correct their own mistakes~\citep{self_refine, shinn2023reflexion}, suggesting a potential way to improve VLMs without additional training. This behavior is termed \emph{self-correction}, a framework that refines LLM responses using LLMs during inference, possibly with external tools or knowledge~\citep{self_debug,gou2024critic}. However, follow-up works in LLMs~\cite{Kamoi2024WhenCL,llm_cannot_self_correct} argue that LLMs struggle to self-correct without access to \emph{oracle feedback}. Currently, there is no clear consensus on when LLMs can effectively perform self-correction~\citep{Kamoi2024WhenCL}. Prior work suggests that self-correction is limited by feedback quality~\citep{gou2024critic,olausson2024is} and is more reliable with tools like search engines or compilers~\citep{llm_cannot_self_correct}.

%\DA{I rewrote a bit this paragraph, put in a new line and joined with the second. I let with \%}
In this work, we explore self-correction in VLMs with a focus on multi-modal understanding, connecting language to visual concepts—a largely unexplored area to date. Specifically, we investigate self-correction within semantic grounding tasks, as illustrated in Fig.\ref{fig:teaser}. Semantic grounding is well-suited for this exploration because it demands the integration of language and visual concepts, requires fine-grained visual understanding, and involves multi-modal reasoning, all of which have significant real-world applications as well as the task itself~\citep{Vasudevan2018ObjectRI,mitchell-etal-2013-generating, deruyttere-etal-2019-talk2car}.
More importantly, VLMs have demonstrated the ability to provide useful feedback in some visual tasks~\citep{lu2024llmscore, zhang2023gpt}, leaving the door open for self-correction in VLMs.
\iclr{
%In this paper, we explore to improve semantic grounding in VLMs through self-correction, as shown in Fig.\ref{fig:teaser}. 
% In this paper, 
Specifically,
% we explore self-correction in the context of semantic grounding, as shown in Fig.\ref{fig:teaser}. 
%\DA{Same thing that above. Consider simply: In this paper, we explore self-correction in the context of semantic grounding.}
%In this paper, we incorporate the semantic grounding in VLMs into the self-correction framework by introducing automated prompt-based feedback, as shown in Fig.\ref{fig:teaser}. 
we focus on two key questions: \textbf{(i)} Can VLMs receive and understand grounding feedback? \textbf{(ii)} Can VLMs provide grounding feedback? %Finally \textbf{(Q3)} can VLMs correct their own mistakes using VLMs during inference.
We then combine the key findings from these two questions to evaluate whether VLMs can self-correct their mistakes by leveraging another instance of the same model during inference.
%\DA{We then combine the key findings from Q1 and Q2 to evaluate whether VLMs can self-correct their mistakes by leveraging another instance of the same model during inference.}
To mitigate the high difficulty of generating reliable feedback, we identify that semantic grounding can be \emph{decomposed} into easier binary verification tasks, therefore, getting more reliable feedback. 

We evaluate the effectiveness of self-correction in our context by repurposing panoptic segmentation datasets from ADE20k~\citep{ade} and COCO~\citep{cocodataset} for semantic grounding~\citep{som_prompt,gptroi}.
%We demonstrate the effectiveness of self-correction framework by repurposing panoptic segmentation datasets from ADE20k~\citep{ade} and COCO~\citep{cocodataset} for semantic grounding~\citep{som_prompt,gptroi}. 
%\DA{For the beginning of this sentence, consider instead: To investigate the effectiveness of self-correction  in our context, we repurpose panoptic segmentation datasets .... }
We analyze three state-of-the-art open-source VLMs (LLaVA-1.5~\cite{llava_1.5}, ViP-LLaVA~\cite{vip_llava}, and CogVLM~\cite{cogvlm}) and two proprietary VLMs (GPT-4V~\cite{gpt4v_report} and GPT-4o) to identify consistent trends. 
Finally, \emph{with no additional finetuning and no access to the oracle feedback}, we show that the self-correction framework improves semantic grounding performance in VLMs by up to 8.4 accuracy points.
}

Below, we summarize the key findings in our exploration:

\iclr{
\textbf{1. VLMs can receive and understand feedback to improve semantic grounding.} With a single round of oracle binary feedback, open-source VLMs improve their semantic grounding performances up to 9 accuracy points, suggesting the feedback potentials to improve grounding performance in VLMs (Sec.~\ref{sec:empirical_receive_feedback}). 

\textbf{2. VLMs can provide high-quality feedback for themselves.} By decomposing semantic grounding into an easier binary verification step and adopting visual prompts, the identified binary verification mechanism improves feedback quality up to an 18-point in $F_1$ score compared to the baseline (Sec.~\ref{sec:empirical_give_feedback}).

\textbf{3. Under the iterative self-correction framework, VLMs improve semantic grounding accuracy up to 8.4 accuracy points \emph{without} the access to the oracle.} Across five VLMs, including three open-source and two proprietary, GPT-4V and GPT-4o, our findings \emph{consistently} indicate that feedback enhances semantic grounding in VLMs (Sec.~\ref{sec:main_results}). 

\textbf{4. Open-source VLMs make errors in semantic grounding even if feedback explicitly states the ground truths.} 
% The fact that some models still fail in 1 out of 4 cases in this scenario indicates a lack of prompt-following capabilities that should be investigated further.
The fact that some models could fail in approximately 25\% of cases in this scenario highlights a deficiency in prompt-following capabilities that should be investigated further (Sec.~\ref{sec:empirical_receive_feedback}).

\textbf{5. Strong proprietary VLMs show significant improvement but still retain limited capability in leveraging ground-truth oracles}. 
% With oracle binary feedback indicating the predictions correctness, strong VLMs like GPT-4V and GPT4o fail to provide correct responses after three turns in more than 30\% of all cases (Sec.~\ref{sec:main_results}).
After three rounds of binary oracle feedback, GPT-4V and GPT-4o improve grounding accuracy substantially but still maintain error rates above 40\% on the ADE20k dataset (Sec.~\ref{sec:main_results}).

%\DA{Lets try to ad also limitations,  otherwise it sounds like we are proposing a technique,the reality is not as rosy as advertised, }

\iffalse
\DA{

1. VLMs still make errors on semantic object identification, even if the ground truth answer is explicitly stated in the model's context (Class label Feedback in Tab 1). The fact that some models still fail in 1 out of 4 cases in this scenario indicates a gigantic lack of prompt-following capabilities that should be investigated further.

2. Similarly, VLMs show limited capability in leveraging ground-truth oracles. Even with multi-turn feedback from the noise-free verification module, strong modules like GPT-4V fail to give the correct reply after 3 turns in almost half of all cases.}
\fi
}

\begin{figure*}[t]
\includegraphics[width=\textwidth]{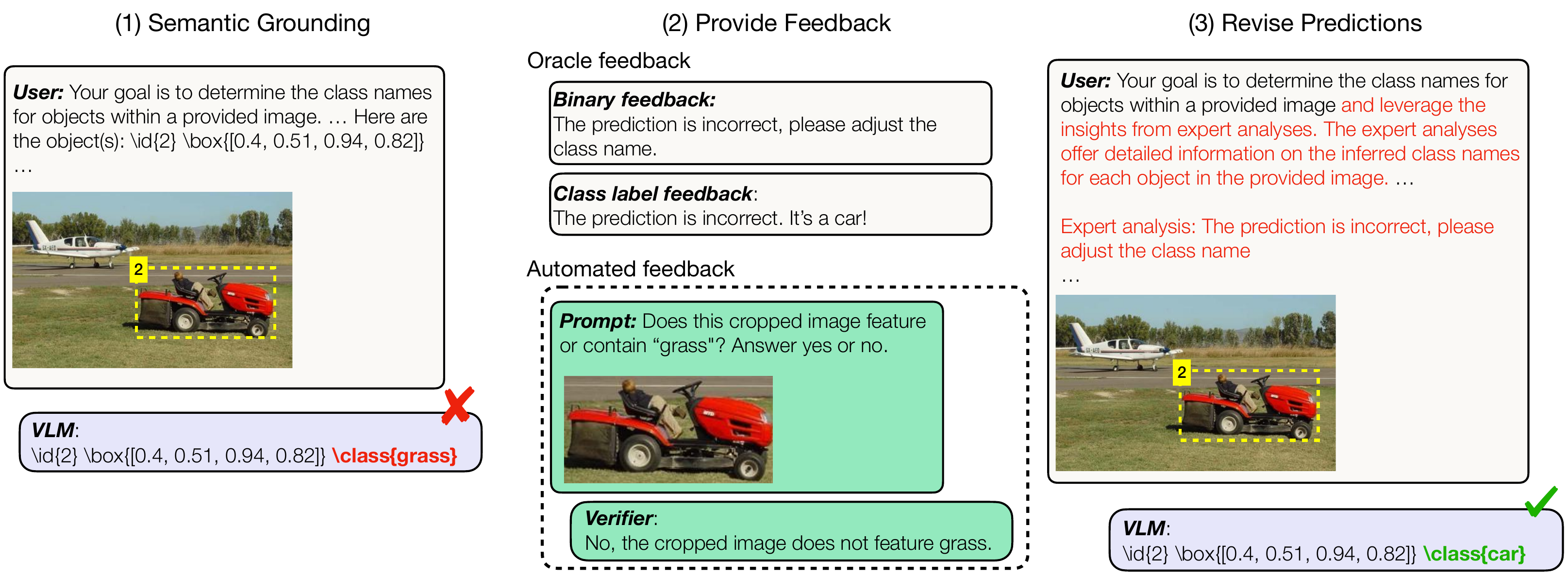}
\centering
\caption{
\iclr{
\textbf{Semantic grounding and self-correction framework.}
}
\textbf{Left} (Semantic Grounding): Given an image and a text prompt that specifies a region of interest, a VLM is tasked to identify the semantic class best describing the image region. 
\textbf{Center} \iclr{(Feedback Generation): For completeness, we explore both oracle and automated feedback generated from VLMs themselves.} \textit{Oracle Binary Feedback}: An oracle provides feedback only on the correctness of the predictions. \textit{Oracle Class Label Feedback}: An oracle provides explicit feedback on the correct class labels. \textit{Automated Binary Feedback}: A VLM acts as a `Verifier', confirms or rejects the previous predictions. %analyzing a modified image with an isolated region of interest to confirm or reject the previous predictions. 
\textbf{Right} (Feedback Integration): VLMs correct their own mistakes by taking the feedback.
}\label{fig:w_and_wo_feedback_demo} 
% \AL{Please review}%\DA{caption needs to be improved an explained}
\end{figure*}

\section{Related Work}
% \subsection{Prompting in LLMs and VLMs}

\iclr{
\textbf{Self-Correction in LLMs:}
LLMs have shown some ability to criticize, refine, and correct their responses through prompt-based feedback~\citep{llm_solve_computer_tasks,self_refine,gou2024critic,li2024confidencemattersrevisitingintrinsic}, supervised finetuning~\citep{havrilla2024glorewhenwhereimprove,zelikman2022star,singh2024beyond} or reinforcement learning~\citep{kumar2024traininglanguagemodelsselfcorrect}. This work examines whether VLMs can self-correct via prompt-based feedback. There remains little consensus on whether LLMs can effectively self-correct through additional prompts~\citep{havrilla2024glorewhenwhereimprove}. While previous studies suggest promise in prompt-based self-correction, they often rely on oracle feedback~\citep{llm_solve_computer_tasks,shinn2023reflexion}, or weak prompts for initial responses~\citep{self_refine,bai2022constitutionalaiharmlessnessai}. Follow-up research suggests that feedback generation limits self-correction~\citep{havrilla2024glorewhenwhereimprove}. 
On the other hand, prompt-based self-correction generally excels when useful external tools, such as code executors or search engines, are accessible~\citep{llm_cannot_self_correct,self_debug,gou2024critic,gao-etal-2023-rarr}; however, these tools are often unavailable in many scenarios.
Fact-checking also shows success, as demonstrated by CoVe, which decomposes generation tasks into simpler verification steps, yielding robust feedback~\citep{dhuliawala2023chainofverificationreduceshallucinationlarge}. Drawing from the extensive literature on LLM self-correction, we analyze whether VLMs can self-correct, focusing on semantic grounding.
}

\iffalse
However, a systematic evaluation of these `self-correction' abilities has shown that the current generation of LLMs can only reliably correct their reasoning responses if given access to external techniques \citep{llm_cannot_self_correct,tyen2024llms}. 
For example, self-debugging~\citep{self_debug} improves code generation by interacting with a code executor, whereas the CRITIC framework uses external tools to cross-check and refine initial responses \citep{gou2024critic}.
This paper presents the first exploration of self-correction for enhancing visual grounding in VLMs. We first verify if VLMs suffer the similar limitations and investigate if visual prompting techniques mitigate this implicit self-bias.
% while exploring the potentials and limitations on VLMs for semantic grounding.
\fi

\textbf{Prompting in LLMs and VLMs:} 
In-context learning in LLMs~\citep{brown2020language} has led to new prompting techniques such as Chain-of-Thought (CoT)~\citep{cot}, Least-to-Most~\citep{ltm}, and StepBack~\citep{stepback} to enhance reasoning capabilities. CoT, in particular, showcases multiple reasoning paths to aid LLMs in solving complex tasks~\citep{tree_of_thoughts, self_consistency_cot}. However, these methods may be less effective in VLMs due to their limited in-context learning~\citep{vlm_icl,zeng2024mllms}. 
%, especially in visually instructed VLMs~\citep{vlm_icl,zeng2024mllms}. 
Conversely, zero-shot CoT promotes model reasoning without the reliance on in-context learning by simply adding a guiding sentence before model responses~\citep{zero_shot_cot}. For VLMs, prompting has predominantly involved visual cues. Studies have shown that models, when trained on extensive web data, can recognize specific visual markers, like red circles~\citep{red_circle}. More recently, Set-of-Marks (SoM) prompting has enabled the GPT-4V to ground multiple objects by overlaying object identifiers on images~\citep{som_prompt, pivot}. Our work incorporates these techniques to provide semantic grounding feedback to VLMs.

\textbf{Multimodal Evaluation and Verification:}
Recent large-scale VLMs like CLIP~\citep{radford2021learning} and GPT-4V~\citep{gpt4v_report} have introduced a new paradigm in multimodal evaluation. For example, traditional metrics struggle to accurately evaluate image captions~\citep{kilickaya-etal-2017-evaluating, Cui_2018_CVPR}. CLIPScore~\citep{clipscore} leverages web-scale VLMs to assess the similarity between images and captions, aligning evaluations more closely with human judgments. Similarly, LLMScore~\citep{lu2024llmscore} combines an image captioner with an off-the-shelf object detector to measure alignment for text-to-image models directly. More recently, GPT-4V has been applied as an automatic evaluator for vision language tasks, such as text-to-3D generation and embodied question answering~\citep{wu2024gpt4vision, OpenEQA2023}. Motivated by the potential of using large VLMs as evaluators, we investigate their capability to evaluate and verify \emph{their own predictions}, marking a shift from earlier approaches that separated predictors from verifiers.

\iclr{\section{Self-Correction in VLMs for Semantic Grounding}\label{sec:setup}}

\iffalse
\AL{

\textbf{Reasons.} Previously, the whole self-correct framework is spread out across three questions, from receiving, generating, to iterative refinements. I feel this confuses the reviewer since we do not provide a clear self-correction framework for them, making it difficult for them to ground each questions into the self-correction framework. 

Rearrange the structures.
\begin{enumerate}
    \item Setup
        \begin{enumerate}
            \item Semantic grounding and notations
            \item Self-Correction and notations
        \end{enumerate}
    \item Questions
    \item Evaluation protocol
\end{enumerate}
}
\fi

In this section, we first define semantic grounding and introduce the adopted self-correction framework for VLMs in Sec.~\ref{sec:semantic_grounding}. We then introduce two key research questions on whether VLMs can correct their own grounding mistakes through self-correction in Sec.~\ref{sec:research_questions}. Finally, we summarize the evaluation metrics, datasets, and models comprising our experiment protocol in Sec.~\ref{sec:exp_protocols}.

\iffalse
In this section, we first define the visual grounding problem for VLMs (Sec.~\ref{sec:semantic_grounding}). 
We then introduce our key research questions on whether VLMs can improve their performances in semantic grounding \new{by} receiving prompt-based feedback and whether they can provide feedback to their own semantic grounding predictions (Sec.~\ref{sec:research_questions}). 
Finally, we summarize the evaluation metrics, datasets, and models comprising our experiment protocol in Sec.~\ref{sec:exp_protocols}.
%\AL{Maybe use a sentence to describe what we mean by feedback dynamics.}
\fi

%\vspace{-1.5mm}

%\subsection{Visual Semantic Grounding}\label{sec:semantic_grounding} 
\subsection{Setup: Semantic Grounding and Self-Correction}\label{sec:semantic_grounding}

\textbf{Semantic Grounding.}
% We consider a visual grounding problem that maps a given image region to the text space, referred to as semantic grounding~\citep{gptroi, som_prompt}. Prior work~\citep{lee2024collavo} has shown that such grounding abilities strongly correlate with the visual reasoning abilities in VLMs.
We study semantic grounding~\citep{gptroi, som_prompt}, mapping image regions to text, which~\cite{lee2024collavo} strongly correlates with visual reasoning abilities in VLMs.
Formally, 
consider an image $\rx \in \R^{\rh \times \rw \times 3}$ where $\rh$ and $\rw$ denote the image's height and width, respectively. There exists a priori image partition function that takes an image and produces $\rN$ semantically distinct regions $\{ \rr_\ri \}_{\ri=1}^\rN$, where each $\rr_\ri \in [0, 1]^{\rh \times \rw}$. 
A general-purpose VLM 
is then tasked to take the image $\rx$, the image region $\rr_\ri$ , a text prompt $\rq$, and to output text $\ro_\ri = \text{VLM}(\rx, \rr_\ri, \rq)$ that best describes the image region. 
The output format depends on the evaluation metrics of interest. 
Fig \ref{fig:w_and_wo_feedback_demo} (left) shows an example task prompt.

Following prior works~\citep{som_prompt,gptroi}, we use ground-truth segmentation masks as semantically distinct image regions $\{ \rr_\ri \}_{\ri=1}^\rN$. We evaluate semantic grounding ability by whether the VLM can estimate the ground-truth class label for each region in every scene.

%\RRM{I think we must be up-front here that we are interested in segmentation and being more specific about class matching ... or we can do it at the end of this section.}

\textbf{Self-Correction.} The term `self-correction' is broadly adopted in LLMs~\citep{Kamoi2024WhenCL}. In this paper, we explore the setup involving explicit feedback generation from the same VLMs. Namely, we use a `Verifier' instantiated from the same VLM to provide feedback on the previous predictions. If feedback suggests further refinement, the VLMs then take the feedback to refine their own predictions. Fig.~\ref{fig:w_and_wo_feedback_demo} highlights the feedback dynamics between VLMs and Verifier.

For an image $x$ and an image region $\rr_i$, we refer the initial predictions without feedback as \textit{base predictions} $\ro_{\ri,0}$. For completeness, we study both oracle feedback $f^*$ and self-generated feedback $f^\text{VLM}$. The feedback can be converted into text or visual marks to help VLMs correct their own mistakes. Appendix~\ref{supp:prompt_tempalte} shows the complete prompt templates.
%We show the examples of each textual and visual prompting technique in Fig.~\ref{fig:textual_and_visual_prompt} (see the full prompt in Appendix~\ref{supp:prompt_tempalte}).

%\subsection{Enhancing Semantic Grounding with Feedback}
%\subsection{Does Feedback Enhancing Semantic Grounding in VLMs?}\label{sec:research_questions}

\subsection{Research Questions}\label{sec:research_questions}

Recently, LLMs have demonstrated significant improvements in performance on complex language semantic tasks such as coding and math reasoning by leveraging %sequential feedback from prompts
self-correction~\citep{self_debug,maf,dhuliawala2023chainofverificationreduceshallucinationlarge,llm_solve_computer_tasks}. 
We note that VLMs can process diverse visual and text inputs while simultaneously sustaining a dialogue from multiple input rounds similar to LLMs. 
To explore whether VLMs behave similarly to LLMs in self-correcting their errors in semantic grounding, we break it into two research questions:

\begin{enumerate}
    \item Can VLMs receive and understand oracle feedback to improve semantic grounding?
    \item Can VLMs provide high-quality binary feedback for themselves? We study binary feedback due to its lower task complexity, leading to a more reliable feedback signal for self-correction. 
\end{enumerate}
By systematically analyzing these two questions, we pave the way to improve semantic grounding in VLMs through self-correction \emph{without the access to oracle feedback} in Section~\ref{sec:iterative_feedback}.
%We study binary feedback due to its lower task complexity, leading to a more reliable feedback signal for self-correction. 
For the rest of this section, we elaborate the questions and setups.
%we direct two series of questions under a noise-free feedback setup: (i) whether VLMs can receive dialogue feedback in order to improve semantic grounding; and (ii) whether VLMs can generate binary grounding feedback for themselves. 

\subsubsection{Can VLMs Receive and Understand \iclr{Oracle} Grounding Feedback?}\label{sec:vlm_receive_feedback}

We start by asking if VLMs can receive and understand oracle grounding feedback to improve base predictions. The results here can provide us an upper bound to improve semantic grounding in VLMs through self-correction. We study this question from two aspects: the types of feedback and the ways to prompt feedback to VLMs.

\textbf{Feedback types.}
We ask: what type of feedback yields the best improvements in grounding performance?
We consider two alternatives: \textbf{(i)} class label feedback -- directly providing the ground-truth class labels in a text prompt; and \textbf{(ii)} binary feedback -- providing a  message on whether the previous prediction is correct.
Fig.~\ref{fig:w_and_wo_feedback_demo} (center) visualizes the two feedback types. 
%See the full prompts in Sec.~\ref{supp:prompt_tempalte}.

\begin{figure}[t]
\includegraphics[width=\columnwidth]{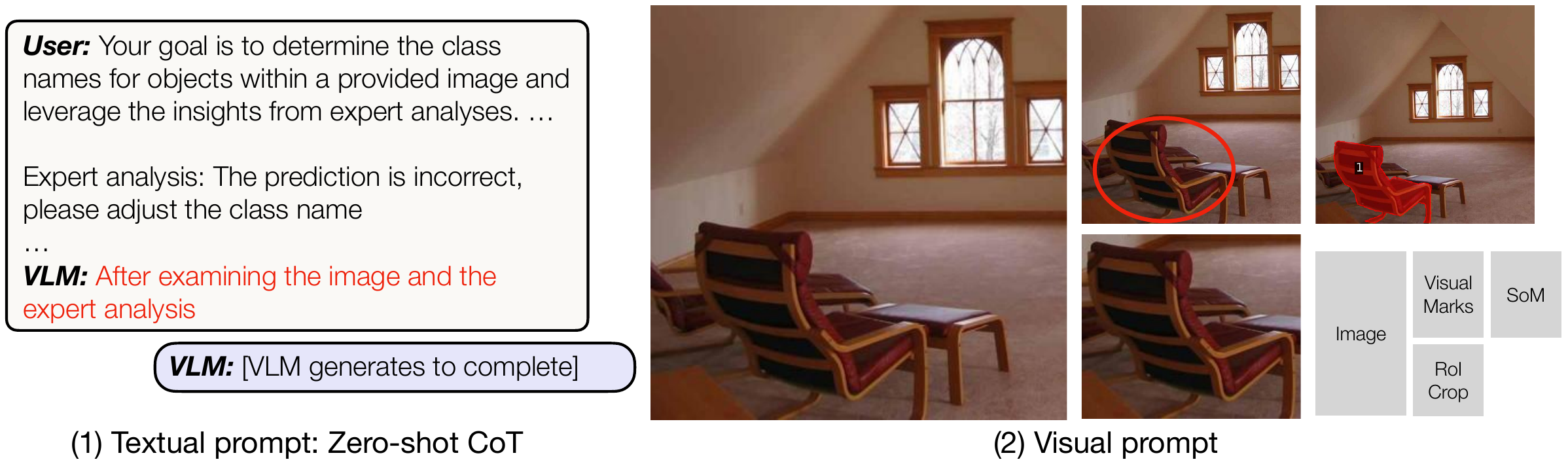}
\centering
\caption{\textbf{Examples of prompting techniques.} 
Left: Zero-shot CoT prepends a guiding sentence (in red) before VLMs' output. % to guide the model. 
Right: We apply various visual prompting techniques including RoI crop, visual marks, and SoM to modify input images to VLMs to guide the models' attention.}\label{fig:textual_and_visual_prompt}
\end{figure}

\iclr{
\textbf{Ways to prompt feedback to VLMs.}
We ask: how should the feedback be prompted to a VLM? 
We consider several alternatives and visualize them in Fig.~\ref{fig:textual_and_visual_prompt}:
}
\textbf{(i)} \textit{Zero-shot Chain-of-Thought (CoT)}: Motivated by~\cite{zero_shot_cot} that shows that simply prepending a guiding sentence \textit{`Let's think step-by-step'} before generation can strongly guide the LLMs for desired tasks, we use the guiding sentence \textit{`After examining the image and the expert analyses, the final answer is [output\_template]'} for the semantic grounding tasks. Here, the feedback is referred as expert analyses to encourage the model to follow the feedback.
\textbf{(ii)} \textit{Visual Marks}: ~\cite{red_circle} shows that Internet-scale vision-language encoders are biased to attend to visual marks (\eg, red circles).
\textbf{(iii)} \textit{Set-of-Mark (SoM)}: ~\cite{som_prompt} shows that overlaying object identifiers on the image improves visual grounding.

%\RRM{Should we have a diagram highlighting an example of each of these?} \DA{+1}

\subsubsection{Can VLMs Give Binary Grounding Feedback for Themselves?} \label{sec:vlm_gives_binary_feedback}

Prior works in LLMs suggest that feedback generation is the bottleneck in self-correction~\citep{gou2024critic,olausson2024is}. The survey paper in LLMs~\citep{Kamoi2024WhenCL} identifies that decomposing complex generation tasks into easier verification tasks enables successful self-correction~\citep{dhuliawala2023chainofverificationreduceshallucinationlarge}. Following this insight, we study binary feedback, a message on whether the previous prediction is correct, due to its lower task complexity. We refer the VLMs performing verification to as `Verifier'. We study binary feedback verification by comparing it with generation-based verification~\citep{self_refine,llm_solve_computer_tasks} referred to as ``intrinsic self-correction'' in prior work~\citep{llm_cannot_self_correct}.
Furthermore, we also study the proper ways to prompt the Verifier.

\textbf{Baseline: intrinsic self-correction.}
We adopt prior work in LLMs self-correction~\citep{llm_solve_computer_tasks} to semantic grounding task. Here, we prompt the verifier to \textit{`Carefully review and refine your answer'} right after the base predictions to automatically correct grounding predictions. 
Although intrinsic self-correction doesn't explicitly generate binary feedback, a binary signal can be obtained by comparing the alignment of grounding predictions before and after correction
%Note that although intrinsic self-correction does not explicitly generate a binary feedback, we can obtain the corresponding binary signal by comparing the alignment of the grounding predictions before and after correction.

\iclr{
\textbf{Ways to prompt the Verifier.}
}
We consider several techniques and visualize them in Fig~\ref{fig:textual_and_visual_prompt}:
\textbf{(i)} \textit{Visual marks:} The verifier receives the image with a highlighted object of interest and a prompt to determine if the predicted class label accurately describes the object \citep{red_circle}.
\textbf{(ii)} \textit{RoI crop:} \new{Prior work~\citep{ViLD}  distills features of cropped regions to the object detectors. Inspired by this, we design the verifier to receive a cropped image isolating the object of interest.}
\textbf{(iii)} A combination of Visual Marks and RoI crop.

\subsection{Experiment Protocols}\label{sec:exp_protocols}

% Our primary experiments focus on panoptic segmentation, which remains a complex visual task containing diverse semantic regions given by segmentation maps. 

\noindent\textbf{Datasets.} 
We analyze the panoptic segmentation dataset from ADE20k~\citep{ade}, which was not previously used for instruction tuning in the open-source VLMs under study. This dataset includes a validation set comprising 2k complex, crowded scenes with over 30k masks across 150 distinct categories. We further validate our results in the iterative setting of COCO panoptic segmentation~\citep{panoptic_seg, cocodataset}. Although the COCO dataset is a standard in visual grounding, most VLMs train on a visual instruction dataset derived from COCO, making it in-domain, unlike ADE20k. The COCO validation set consists of 5k images. Consistent with previous VLM grounding research~\citep{som_prompt}, we utilize the same subset of 100 images from both ADE20k and COCO for our analysis.
\iffalse
We use the panoptic segmentation dataset in ADE20k~\citep{ade} for our analysis.
This dataset was not previously used in the instruction tuning of the open-source VLMs that we study, which allows us to better characterize in-the-wild generalization abilities. 
Further, this dataset contains a validation set of 2k complex, crowded scenes with over 30k masks, labeled across a fine-grained class label spectrum of 150 distinct categories. 
We validate our results on the \new{iterative setting} 
in COCO panoptic segmentation~\citep{panoptic_seg, cocodataset}. Although the COCO dataset is standard in the visual grounding community, most VLMs are trained on the visual instruction dataset derived from the COCO labels, meaning it can be considered as in-domain for most VLMs, in contrast to ADE20k. The COCO validation set has 5k images. 
Consistent with prior VLM grounding works~\citep{som_prompt}, we use the same subset of 100 images for ADE20k and COCO for our analysis. 
\fi
%\\[0.05em]

\noindent\textbf{VLMs.} 
%\subsubsection{VLMs}
%\noindent \textbf{VLMs.} 
We analyze three state-of-the-art open-source VLMs including LLaVA-1.5~\citep{llava_1.5}, ViP-LLaVA~\citep{vip_llava} and CogVLM~\citep{cogvlm}. 
LLaVA-1.5 is a successor of LLaVA~\citep{llava}, a visual instruction tuned VLM, and has scaled up to a larger model and a larger training dataset.
ViP-LLaVA shares the overall model architecture and training strategy with LLaVA, but focuses on synthesizing a diverse set of visual marks in the training dataset, 
effectively improving the model performance when using 
% to encourage a more user-friendly interface for 
visual prompts and allowing for a more user-friendly interface.
CogVLM is a generalist VLM with highlights on integrating image and text features without sacrificing any performance on NLP tasks. %\footnote{We adopt LLaVA-1.5 13b (from \url{https://huggingface.co/llava-hf/llava-1.5-13b-hf}), ViP-LLaVA 13b (from \url{https://huggingface.co/llava-hf/vip-llava-13b-hf}), and CogVLM (from \url{https://huggingface.co/THUDM/CogVLM})}
%\neurips{We include all the prompt templates in the Appendix~\ref{supp:prompt_tempalte}.}

%Each VLM receives as input, the image and the ground truth segmentation maps $g(x)$.
%\\[0.05em]

%\subsubsection{Metrics.}
\noindent \textbf{Grounding metrics.} 
We evaluate semantic grounding performance by measuring classification accuracy. We use off-the-shelf sentence embeddings~\citep{all-mpnet-base-v2} to map the VLM outputs $o_i$ to the label from the class label list with the largest cosine similarity. We then report accuracy aggregated over all regions $\rr_{\ri}$ for each scene in the dataset. 
While it is not perfect, our quantitative analysis in Appendix~\ref{supp:class_mapping} demonstrates that the errors are within a reasonable range.

\noindent \textbf{Feedback metrics.} We assess the VLM verifier's capability to generate a binary feedback signal in $F_1$ scores, considering the imbalanced distribution of oracle binary feedback.  Appendix~\ref{supp:feedback_accuracy} shows that $F_1$ is a more representative metric than accuracy when evaluating feedback quality.

\begin{table}[t]
\centering
\resizebox{\columnwidth}{!}{%
\begin{tabular}{@{}lrrr@{}}
\toprule
                          & LLaVA-1.5 & ViP-LLaVA & CogVLM \\ \midrule
\rowcolor[HTML]{EFEFEF} 
Base prediction           & 35.86     & 35.86     & 15.98  \\
~~ + Binary Feedback      & 41.04     & 40.36     & 16.25  \\
~~ + Class Label Feedback & 94.8      & 74.99     & 77.04  \\ \bottomrule
\end{tabular}%
}
\caption{\textbf{Findings of different feedback types.} VLMs can receive different types of oracle feedback to improve grounding accuracy with no additional prompting techniques (\ie zero-shot CoT or visual prompts). Among the evaluated VLMs, LLaVA-1.5 has the largest gains, improving 5.18 and 61.06 when provided with oracle binary and class label feedback, respectively.}
\label{tab:oracle_feedback_types}
\end{table}

\begin{table}[t]
\centering
\resizebox{\columnwidth}{!}{%
\begin{tabular}{@{}lrrrrr@{}}
\toprule
                                       & Zero-shot CoT & Visual Prompt & LLaVA-1.5 & ViP-LLaVA & CogVLM \\ \midrule
\rowcolor[HTML]{EFEFEF} 
Base prediction                        & No            & No            & 35.86     & 35.86     & 15.98  \\ \cmidrule(l){2-6} 
                                       & No            & No            & 41.04     & 40.36     & 16.25  \\
                                       & Yes           & No            & 43.3      & 42        & 18.25  \\
                                       & Yes           & SoM           & 42.41     & 44.53     & 18.64  \\
\multirow{-4}{*}{~~ + Binary Feedback} & Yes           & Visual marks  & \textbf{45.38}& \textbf{45.21}& \textbf{19.46}\\ \bottomrule
\end{tabular}%
}
\caption{\textbf{Findings of ways to prompt binary feedback to VLMs.} We find that combining zero-shot and visual marks leads to an average of 7.45 gains \wrt the base predictions.}
\label{tab:oracle_binary_feedback}
\end{table}

\section{Empirical Findings}\label{sec:empirical_findings}

In this section, we experiment on ADE20k dataset to study the questions in Sec.~\ref{sec:research_questions}.
All experiments are run on three different seeds and we report the average performances. 
% except for the experiments involving proprietary VLMs, \eg GPT-4V~\citep{gpt4v_report}.
%\AL{TODO: expand each empirical findings more}

\subsection{Can VLMs Receive and Understand \iclr{Oracle} Grounding Feedback?}\label{sec:empirical_receive_feedback}  %\DA{nitpicking. I find the word receive alone a tiny bit weird, may be we want to say receive and understand feedback. }

Table~\ref{tab:oracle_feedback_types} and Table~\ref{tab:oracle_binary_feedback} summarize the base predictions for each model and the improved grounding accuracies after receiving \iclr{oracle} grounding feedback.

\textbf{Findings of feedback types.}
With no additional prompting techniques (\ie, zero-shot CoT or visual prompts), Table~\ref{tab:oracle_feedback_types} demonstrates that LLaVA-1.5 has the largest gains among the evaluated VLMs. When provided with oracle binary and class label feedback, LLaVA-1.5 improves the grounding accuracies by 5.18 and 61.06 points, respectively. We find that VLMs can receive and understand oracle feedback to improve performance, without requiring any additional data or architectural modifications. 

Intuitively, oracle class label feedback yields the most improvement, since it directly reveals the class labels and consequently reduces the semantic grounding task to a text retrieval problem. Perhaps surprisingly, oracle class label feedback does not automatically improve accuracy to 100\%. 
% This outcome points to a limitation in the ability of open-source VLMs to handle tasks relying purely on language understanding, suggesting an area for potential improvement in these models~\citep{lin2023vila}. \iclr{
% The fact that some models still fail in 1 out of 4 cases in this scenario indicates a gigantic lack of prompt-following capabilities that should be investigated further.
This outcome highlights a limitation in open-source VLMs' ability to perform tasks based solely on language understanding, indicating a potential area for improvement in these models~\citep{lin2023vila}. Indeed, some models fail in approximately 25\% of cases in this scenario, demonstrating a significant deficiency in prompt-following capabilities that warrants further investigation (see Table~\ref{tab:oracle_feedback_types}, Class Label Feedback)

\textbf{Findings of ways to prompt binary feedback to VLMs.}
Given the simplicity and effectiveness of oracle binary feedback, we further investigate the ways to prompt binary feedback to VLMs. In Table~\ref{tab:oracle_binary_feedback}, we explore both textual and visual prompts to encourage the uses of the oracle binary feedback.

We first show that zero-shot CoT augments \iclr{oracle} binary feedback for every model considered by up to 2.26 accuracy points. 
This aligns with trends in LLMs that suggest the effectiveness of CoT to improve reasoning~\citep{cot,zero_shot_cot}. 
On the other hand, visual prompting with SoM \citep{som_prompt} does not significantly improve beyond zero-shot CoT for models that were not already pre-trained with data featuring visual prompting cues (\eg, LLaVa-1.5).
In contrast, ViP-LLaVA was specifically trained for interpreting visual cues; this model improves with both SoM and visual marks (\eg, red circles).
Notably, the combination of zero-shot CoT and visual marks emerges as the most effective strategy, increasing by \cvpr{an average of} 7.45 grounding accuracy points relative to the base predictions.
Thus, for open-source VLMs, we identify that the best way to introduce binary feedback in semantic grounding is to combine visual marks and zero-shot CoT.

\begin{table}[t]
\centering
\resizebox{\columnwidth}{!}{%
\begin{tabular}{@{}lrrrrr@{}}
\toprule
%\multicolumn{1}{c}{} & \multicolumn{1}{c}{Visual prompt} &  \multicolumn{1}{c}{Zero-shot self-consistency} & \multicolumn{1}{c}{LLaVA-1.5} & \multicolumn{1}{c}{Vip-LLaVA}  & \multicolumn{1}{c}{CogAgent} \\ \midrule
\multicolumn{1}{l}{} & \multicolumn{1}{r}{Visual prompt}  & \multicolumn{1}{c}{LLaVA-1.5} & \multicolumn{1}{c}{ViP-LLaVA}  & \multicolumn{1}{c}{CogVLM} \\ \midrule
\rowcolor[HTML]{EFEFEF} Intrinsic Self-Correction                   & N/A                   & 51.12                        & 48.19                          &                              21.87\\ 
%VLMs w/ text prompt & No                 & 36.9  &        58.18                        &     36.73                         \\
\cmidrule(l){2-5} 
\multirow{3}{*}{VLM Binary Verification}& Visual marks           & 56.16                        &                                \textbf{60.47}&                              39.16\\
%                                  & Red circle           & Yes                                         &                              45$^\dagger$&                                53.72&                               37.61\\
                                  & RoI crop                 & \textbf{61.71}                        &                                58.18&                              \textbf{40.68}\\
%                                  & Crop                 & Yes                                         &                              34.81&                                23.08&                              38.6\\
                                  & Visual marks + RoI crop    &                         61.14&                                59.6&                              39.79\\  \bottomrule
%                                  & Crop + Red circle    & Yes                                         &                              43.77$^\dagger$&                                48.55&                                38.61\\ \bottomrule
\end{tabular}%
}
\caption{
\textbf{VLMs can provide high-quality grounding feedback for themselves (higher $F_1$ scores).}
Compared to intrinsic self-correction, VLM binary verification achieves at least 10 points higher in feedback quality. Additionally, we find that visual prompting techniques should be tailored to the specific VLMs, \eg, ViP-LLaVA prefers visual marks. We bold the best performances of each VLM.}
\label{tab:provide_feedback_f1}
\vspace{-3mm}
\end{table}

\subsection{Can VLMs Give Binary Grounding Feedback for Themselves?}\label{sec:empirical_give_feedback}

We assess the quality of binary feedback using $F_1$ scores due to potentially imbalanced oracle feedback. Table~\ref{tab:provide_feedback_f1} provides $F_1$ scores of intrinsic self-correction and the binary feedback produced by a VLM Verifier.

\iclr{
\textbf{Results.}
}
We first assess the effectiveness of intrinsic self-correction, which involves continuing another round of conversation by asking \textit{`Carefully review and refine your answer'} to the VLM and directly outputting the revised predictions. We derive the binary feedback by comparing whether the revised predictions differ from the initial predictions.
When evaluated in accuracy, intrinsic self-correction achieves low accuracies at 47.03, 47.13, and 59.5 on LLaVA-1.5, ViP-LLaVA, and CogVLM, respectively.
Aligned with the previous studies on LLMs, we find that VLMs struggle to improve via intrinsic self-correction out-of-the-box.
%The VLMs results here are aligned with previous studies on LLMs~\cite{Kamoi2024WhenCL} that LLMs struggle to improve via intrinsic self-correction out-of-the-box.

In Table \ref{tab:provide_feedback_f1}, we identify that binary verification mechanism for VLM using RoI crop significantly improves the $F_1$ score for all three models, by up to 18.81 points. This observation aligns well with the strong self-evaluation capabilities in LLMs. We may also augment this binary verification with visual marks such as red circles. Additionally, the choice of visual prompting technique should be tailored to the specific VLM. For instance, RoI crop tends to be more effective for networks not trained on visual marks (\eg, LLaVA-1.5 and CogVLM), while visual marks yield better results for models accustomed to such cues (\eg, ViP-LLaVA).

\begin{table*}[t]
\centering
\resizebox{0.8\textwidth}{!}{%
\begin{tabular}{@{}lrrrrrrr@{}}
\toprule
\multirow{2}{*}{VLM}       & \multirow{2}{*}{Binary feedback source} & \multicolumn{6}{c}{Dialogue round}                                                                                                                        \\ \cmidrule(l){3-8} 
                           &                                  & \multicolumn{1}{c}{$\rt=0$} & \multicolumn{1}{c}{$\rt=1$} & \multicolumn{1}{c}{$\rt=2$} & \multicolumn{1}{c}{$\rt=3$} & \multicolumn{1}{c}{$\rt=4$} & \multicolumn{1}{c}{$\rt=5$} \\ \midrule
%\rowcolor[HTML]{EFEFEF} OpenSeeD                    & \cmark & N/A                               & ???                & -                       & -                       & -                       & -                       & -                       \\ \cmidrule{2-9} 
                           & Intrinsic Self-Correction        & 35.86                   & {\color[HTML]{CB0000} 30.92}& {\color[HTML]{CB0000} 29.64}& {\color[HTML]{CB0000} 28.54$_{-7.32}$}& -                       & -                       \\
                           & VLM Verification \small{\textbf{(ours)}}                   & 35.86                   & 37.97                   & 38.93                   & 39.27                   & 39.54                   & 40.29$_{+4.43}$                   \\
\multirow{-3}{*}{LLaVA-1.5} & Oracle Verification \small{\textbf{(ours)}}                       & 35.86                   & 45.42                   & 47.95                   & 51.55                   & 52.04                   & 53.2$_{+17.34}$                    \\ \cmidrule(l){2-8} 
                           & Intrinsic Self-Correction        & 35.86                   & {\color[HTML]{CB0000} 27.72}& {\color[HTML]{CB0000} 26.7}& {\color[HTML]{CB0000} 25.68$_{-10.18}$}& -                       & -                       \\
                           & VLM Verification \small{\textbf{(ours)}}                   & 35.86                   & {\color[HTML]{CB0000} 35.14}& 36.06                   & 36.37                   & 36.16& 36.47$_{+0.39}$                   \\
\multirow{-3}{*}{ViP-LLaVA} & Oracle Verification \small{\textbf{(ours)}}                       & 35.86                   & 47.45                   & 47.64                   & 50.54                   & 51.82                   & 53.13$_{+17.27}$                   \\ \cmidrule(l){2-8} 
                           & Intrinsic Self-Correction        & 15.98                   & {\color[HTML]{CB0000} 8.33}& {\color[HTML]{CB0000} 8.6}& {\color[HTML]{CB0000} 9.08$_{-6.9}$}& -                       & -                       \\
                           & VLM Verification \small{\textbf{(ours)}}                   & 15.98                   & 17.13                   & 17.96                   & 18.09                   & 18.5                    & 18.64$_{+2.66}$                   \\
\multirow{-3}{*}{CogVLM}&   Oracle Verification \small{\textbf{(ours)}}                       & 15.98                   & 19.6                    & 20.96                   & 21.51                   & 21.82                   & 22.12$_{+6.14}$                   \\ \cmidrule{2-8} 
% \rowcolor[HTML]{EFEFEF} GPT-4V \& SoM                    &\cmark & N/A                               & $63.4^*$                & -                       & -                       & -                       & -                       & -                       \\ \bottomrule
                           & Intrinsic Self-Correction        & 40.36& {\color[HTML]{CB0000} 22.33}& {\color[HTML]{CB0000} 25.2}& {\color[HTML]{CB0000} 22.95$_{-17.41}$}& -                       & -                       \\
                           & VLM Verification \small{\textbf{(ours)}}                   & 40.36& 41.8& 43.23& 42.4$_{+2.04}$& -& -\\
\multirow{-3}{*}{GPT-4V} & Oracle Verification \small{\textbf{(ours)}}                       & 40.36& 50& 52.45& 53.27$_{+12.91}$& -& -\\ \cmidrule{2-8}  
% \rowcolor[HTML]{EFEFEF} GPT-4V \& SoM                    &\cmark & N/A                               & $63.4^*$                & -                       & -                       & -                       & -                       & -                       \\ \bottomrule
                           & Intrinsic Self-Correction        & 33.81& 34.01& 39.13& 37.5$_{+3.68}$& -                       & -                       \\
                           & VLM Verification \small{\textbf{(ours)}}                   & 33.81& 39.13& 40.98& 41.18$_{+7.36}$& -& -\\
\multirow{-3}{*}{GPT-4o} & Oracle Verification \small{\textbf{(ours)}}                       & 33.81& 49.59 & 54.91&  57.78$_{+23.91}$& -& -\\ \bottomrule
\end{tabular}%
}\vspace{1mm}
\caption{\textbf{Performances of self-correction on ADE20k up to 5 rounds.} Oracle binary feedback (oracle verification) consistently improves all evaluated VLMs. Without oracle, VLM verification \emph{still} consistently improve the grounding performances. On the other hand, intrinsic self-correction demonstrates negative gains in almost every VLM except for GPT-4o. We, therefore, stop intrinsic self-correction in the third round. For GPT-4V and GPT-4o, we find that running for three rounds is enough to identify the consistent trend.
Red-colored font indicates the performances is lower than performances when $\rt=0$. Numbers in the subscript indicate the performance changed \wrt to the performances when $\rt=0$.
%For the iterative approaches, we highlight the performance difference \wrt its performance at $\rt=0$. We mark the non-iterative approach in grey background and use red-colored font to indicate performance decrease \wrt the base predictions. 
%\AL{ToDo: Add OpenSeeD}
%\DA{mark decrease in performance with red, increase in performance with green , consider moving GPT to the bottom of the table} 
%\RRM{Also need to change `VLM Binary Verification' to new name} \RRM{We can also include (ours) for our method} 
}
%\DA{a common issue reviewers have was the inconsistency in t for some VLMs/method, ideally we should hv them all at the same t, if no time for the experiment, let's put in the caption why}
\label{tab:iteratively_improve_ade}
\vspace{-3mm}
\end{table*}

\begin{table*}[t]
\centering
\resizebox{0.8\textwidth}{!}{%
\begin{tabular}{@{}lrrrrrrr@{}}
\toprule
\multirow{2}{*}{VLM}                             & \multirow{2}{*}{Binary feedback source}                         & \multicolumn{6}{c}{Dialogue round}                                                                                                                        \\ \cmidrule(l){3-8} 
    &   & \multicolumn{1}{c}{$\rt=0$} & \multicolumn{1}{c}{$\rt=1$} & \multicolumn{1}{c}{$\rt=2$} & \multicolumn{1}{c}{$\rt=3$} & \multicolumn{1}{c}{$\rt=4$} & \multicolumn{1}{c}{$\rt=5$} \\ \midrule
%\rowcolor[HTML]{EFEFEF} MaskDINO                    & \xmark& N/A                               & ???                & -                       & -                       & -                       & -                       & -                       \\ \cmidrule(l){2-9} 
                            & Intrinsic Self-Correction         & 36.3                    & {\color[HTML]{CB0000} 33.69}& {\color[HTML]{CB0000} 32.26}& {\color[HTML]{CB0000} 31.63$_{-4.66}$}& -                       & -                       \\
                            & VLM Verification \small{\textbf{(ours)}}                    & 36.3                    & {\color[HTML]{CB0000} 35.87}& 36.94                   & 37.04                   & 37.69                   & 38.21$_{+1.91}$                   \\
\multirow{-3}{*}{LLaVA-1.5} & Oracle Verification \small{\textbf{(ours)}}                        & 36.3                    & 41.55                  & 43.81                   & 46.22                   & 47.55                   & 48.77$_{+12.47}$                   \\ \cmidrule(l){2-8} 
                            & Intrinsic Self-Correction         & 37.26                   & {\color[HTML]{CB0000} 32.64}& {\color[HTML]{CB0000} 32.4}& {\color[HTML]{CB0000} 31.12$_{-6.13}$}& -                       & -                       \\
                            & VLM Verification \small{\textbf{(ours)}}                    & 37.26                   & 37.84                   & 39.64                   & 39.64                   & 40.01                   & 40.44$_{+3.18}$                   \\
\multirow{-3}{*}{ViP-LLaVA} & Oracle Verification \small{\textbf{(ours)}}                        & 37.26                   & 44.9                    & 48.08                   & 50.15                   & 51.75                   & 52.54$_{+15.28}$                   \\ \cmidrule(l){2-8} 
                        & Intrinsic Self-Correction         & 14.8                    & 16.23                   & 16.47                   & 15.92$_{+1.11}$                   & -                       & -                       \\
                            & VLM Verification \small{\textbf{(ours)}}                    & 14.8                    & 16.97                   & 17.83                   & 18.3                    & 18.52                   & 18.84$_{+4.04}$                   \\
\multirow{-3}{*}{CogVLM}  & Oracle Verification \small{\textbf{(ours)}}                        & 14.8                    & 19.42                   & 20.14                   & 20.7                    & 21.01                   & 21.25$_{+6.45}$                   \\ \cmidrule(l){2-8} 
%\rowcolor[HTML]{EFEFEF} GPT-4V \& SoM                   & \cmark& N/A                               & $75.7^*$                & -                       & -                       & -                       & -                       & -                       \\  \bottomrule
                            & Intrinsic Self-Correction         & 40.92& {\color[HTML]{CB0000} 30.89}& {\color[HTML]{CB0000} 36.62}& {\color[HTML]{CB0000} 32.8$_{-8.12}$}& -& -\\
                            & VLM Verification \small{\textbf{(ours)}}                    & 40.92& 43.94& 44.9& 45.38$_{+4.46}$& -& -\\
\multirow{-3}{*}{GPT-4V}  & Oracle Verification \small{\textbf{(ours)}}                        & 40.92& 52.7& 56.5& 57.8$_{+16.88}$& -& -\\ \cmidrule(l){2-8} 
%\rowcolor[HTML]{EFEFEF} GPT-4V \& SoM                   & \cmark& N/A                               & $75.7^*$                & -                       & -                       & -                       & -                       & -                       \\  \bottomrule
                            & Intrinsic Self-Correction                                          & 39.49 & 47.13 & 48.08 & 46.65$_{+7.15}$& -& -\\
                            & VLM Verification \small{\textbf{(ours)}}                    & 39.49 & 46.49& 47.77& 47.92$_{+8.43}$& -& -\\
\multirow{-3}{*}{GPT-4o}  & Oracle Verification \small{\textbf{(ours)}}               & 39.49 & 57& 62.26& 67.19$_{+27.69}$& -& -\\ \bottomrule
\end{tabular}%
}\vspace{1mm}
\caption{\textbf{Performances of self-correction on COCO up to 5 rounds.} Oracle binary feedback (oracle verification) consistently improves all evaluated VLMs. Without oracle, VLM verification \emph{still} consistently improve the grounding performances. On the other hand, intrinsic self-correction demonstrates negative gains in almost every VLM except for GPT-4o and CogVLM. We, therefore, stop intrinsic self-correction in the third round. For GPT-4V and GPT-4o, we find that running for three rounds is enough to identify the consistent trend.
Red-colored font indicates the performances is lower than performances when $\rt=0$. Numbers in the subscript indicate the performance changed \wrt to the performances when $\rt=0$.
}
%\vspace{-5mm}
\label{tab:iteratively_improve_coco}
\vspace{-3mm}
\end{table*}

%\section{Towards Iterative Self-Generated Feedback for Semantic Grounding in VLMs}

%\section{Towards Iterative Self-Generated Feedback for Semantic Grounding}
\iclr{\section{Can VLMs Correct their Grounding Errors through Self-Correction?}}
\label{sec:iterative_feedback}

%We have established that VLMs can improve on semantic grounding when given (noise-free) feedback. Furthermore, VLMs can also act as Verifiers that produce poentially noisy, binary feedback. 

Our key findings in Sec.~\ref{sec:empirical_findings} show that \textbf{(1)} VLMs can receive and understand oracle feedback and \textbf{(2)} VLMs can give binary feedback for themselves. We now combine them to evaluate whether VLMs can self-correct their mistakes by leveraging another instance of the same model. Furthermore, can VLMs \emph{iteratively} perform self-correction to trade compute for performances?

\subsection{Setup: Iterative Self-Correction in VLMs}

%\subsection{Binary \emph{Self-Feedback} in VLMs}

We introduce an iterative dialogue loop between a VLM agent and Verifier, where at the first timestep $\rt=0$, the VLM obtains base predictions $\{\ro_{\ri, 0}\}_{\ri=1}^\rN$ for every scene (Sec. \ref{sec:semantic_grounding}). We then prompt the Verifier to generate a binary feedback signal for every prediction $f^\text{VLM}(\rx, \rr_\ri, \ro_{\ri, 0})$ (Sec.~\ref{sec:vlm_gives_binary_feedback}). In the next timestep, the VLM agent uses this binary feedback to revise predictions (Sec.~\ref{sec:vlm_receive_feedback}). We repeat these steps to a maximum iteration count or until the verifier agrees with the prediction. 
%Fig.~\ref{fig:qualitative_res} demonstrates the iterative interactions between a VLM agent and the Verifier.
%with the predictive binary feedback, we run the VLMs along with zero-shot CoT and visual prompt described in Sec.~\ref{sec:research_questions}. Finally, we iterate the last two steps until converge.

\cvpr{In our experiments, to encourage feedback receiving, we use the textual prompts (\ie, zero-shot CoT) and the visual prompts (\ie, red circles for open-source VLMs and SoM for proprietary VLMs).
To generate binary feedback, we use RoI crop. }
Consistent with prior work~\citep{som_prompt}, we use the same subset of 100 images for ADE20k and COCO for our analysis.
%We report the experimental results on ADE20k and COCO. 
%We follow \citep{som_prompt} and implement our framework on a subset of ADE20k and COCO dataset, which we call ADE20k-SoM and COCO-SoM.
%We include full dataset details in the Appendix~\ref{supp:dataset_details}.
%We apply the textual and visual prompts to all the iterative feedback approaches. 
%The full table with different configurations is given in the Appendix.
%\\[0.01em]

\noindent \textbf{Baselines.} 
% \DA{I wouldnt call this here baselines, too risky, but i am unsure how to call it now..}
% \noindent \textbf{Approaches for Reference} \DA{I wouldnt call this here baselines,  , but i am unsure how to call it , leaving this name for now..} \RRM{I'm okay with baselines. We could also use "Alternatives", or "Comparisons", or "Benchmarks"} \AL{why not baselines? because we didn't implement it by ourselves?}
%\DA{I think this should go up, it should be fine to say we used those only in this section and the main study was done in ADE.}
%To provide a complete analysis, we introduce COCO dataset as another panoptic segmentation dataset~cite{coco}. While COCO dataset is more standard to the visual grounding community, most VLMs are trained on the visual instruction dataset derived from COCO labels. We, therefore, consider it as in-domain dataset, as compared to ADE20k-150~\citep{ade}. The COCO validation set is composed of 5000 images. Consistent with prior works on VLM grounding~\citep{som_prompt}, we use a subset of 100 random images for our analysis. For VLMs, we include the current SOTA model, LLaVA-1.6~\citep{llava_1.6}, with improved reasoning, OCR and world knowledge as compared to LLaVA-1.5~\citep{llava_1.5}. 
Similar to Sec.~\ref{sec:vlm_gives_binary_feedback}, we adopt intrinsic self-correction adopted~\citep{llm_solve_computer_tasks} as our baseline.
To identify the self-correction upper bounds of each VLM, we also report the performances of self-correction with the access to oracle binary feedback, referred to as Oracle Verification. 

\iffalse
We compare our VLM binary verification approach with an alternative feedback-based baseline that is inspired by prior work in LLMs~\citep{llm_solve_computer_tasks,llm_cannot_self_correct}.
\neurips{We refer to it as intrinsic self-correction.}
% and uses only intrinsic self-correction. 
We also compare to a Noise-Free approach, where the feedback is provided by an oracle and assumed to be perfect; the latter acts as an upper bound achievable only with access to an oracle. 
\fi

\noindent \textbf{Proprietary VLMs.} 
Open-source VLMs often suffer from shorter context window or limited instruction following capabilities. We, therefore, experiment the identified self-correction framework using GPT-4V~\citep{gpt4v_report} and %its successor 
GPT-4o.

\noindent \textbf{Base predictions generation.} 
The self-correction survey in LLMs~\citep{Kamoi2024WhenCL} finds that the weak initial predictions can lead to false promises in self-correction.
We attempted to improve open-source VLMs by adding SoM prompt, but observed significant performance drops compared to using bounding boxes alone. For LLaVA-1.5, the base predictions achieve 35.86 in ADE20k. However, adding SoM and using RoI crop result in 11.06 and 19.67, respectively.
This may be because most open-source VLMs, including the three in our study, are trained to identify image regions using bounding boxes~\citep{gptroi,you2023ferret}.
In contrast, proprietary VLMs have shown strong improvements with SoM~\citep{som_prompt}. Therefore, we adopted SoM to generate base predictions for GPT-4V and GPT-4o.

\subsection{Main Results}\label{sec:main_results}

%\AL{consider add GPT-4V discussion here and kindly mention that we use ``OpenAI API for inference and perform the exact same evaluation as described in Sec~\ref{sec:exp_protocols}''.}
% Tables~\ref{tab:iteratively_improve_ade} and~\ref{tab:iteratively_improve_coco} present iterative feedback results for the ADE20k and COCO datasets. 
% In ADE20k (Table~\ref{tab:iteratively_improve_ade}),
\textbf{Open-source VLMs.}
Tables~\ref{tab:iteratively_improve_ade} and~\ref{tab:iteratively_improve_coco} illustrate that multiple rounds of oracle binary feedback consistently enhance the performance of all open-source VLMs, with gains ranging from 6.14 to 17.34 in ADE20k and 6.45 to 15.28 in COCO. 
%Additionally, multiple self-correction increase grounding accuracy by up to 7.78 and 7.64 points on ADE20k and COCO, respectively, compared to a single round (\ie, $\rt=1$). 
The identified VLM binary verification, despite producing noisy feedback, also \emph{consistently} improves grounding accuracy by 0.39 to 4.43 points in ADE20k and 1.91 to 4.04 points in COCO. These gains are consistent across all three open-source VLMs, underscoring the benefits of iterative feedback for zero-shot improvements in grounding accuracy, even with noisy feedback.

In sharp contrast, intrinsic self-correction decreases downstream grounding in all settings by up to 10 points, except where base predictions are weak, such as with CogVLM in COCO. We empirically find that self-correction cannot reliably identify the alreadily correct predictions. \cvpr{We stop intrinsic self-correction in the third rounds due to its consistent negative impacts.}

\textbf{GPT-4V and GPT-4o.}
GPT-4V and GPT-4o improve predictions with both VLM binary feedback and oracle binary feedback, even more than the open-source VLMs do. In particular, GPT-4o significantly improves and sometimes surpasses GPT-4V, especially when incorporating oracle binary feedback. However, perhaps surprising, even with oracle binary feedback indicating prediction correctness, strong GPT-4V and GPT4o fail to provide correct responses after three turns with less than $60$ points accuracy overall in ADE20k. We find that running for three rounds is enough to identify the consistent trend.

Similar to open-source VLMs, GPT-4V exhibit negative gains for intrinsic self-correction. Intriguingly, there are stark differences between GPT-4V and GPT-4o: GPT-4V shows a 17-point drop in accuracy in ADE20k, while GPT-4o sees a 7-point increase in COCO. The reasons for these sharp differences remain unclear due to unknown model architectures and specific training data used in proprietary models. However, we note that the identified VLM binary verification consistently improves upon both base predictions and intrinsic self-correction with enough dialogue rounds.

We emphasize that the identified VLM binary feedback verification requires no access to external tool or oracle.
Thus, our results show that \emph{VLMs can iteratively self-correct their own grounding mistakes when prompted in a proper way.} We anticipate the improvements from iterative self-correction will improve with future VLMs.

\section{Conclusion}\label{sec:conclusion}

This work explores the potentials of self-correction in VLMs in the context of semantic grounding. We break down this into two key research questions \textbf{(i)} Can VLMs receive and understand oracle grounding feedback and \textbf{(ii)} Can VLMs provide grounding feedback? 
%Throughout our systematic analysis, we find that the answers to both questions are positive when prompted in a proper way.
Across two datasets and five VLMs including proprietary ones, we demonstrate that with the identified VLM binary feedback verification, VLMs \emph{can} iterative self-correct their own grounding mistakes. 
Within five rounds of VLM binary feedback, open-source VLMs and proprietary VLMs improve up to 4 and 8 accuracy points. We highlight that the self-correction in VLMs requires \emph{no access to oracle or any finetuning or architectural changes.}

\noindent \textbf{Limitations.} 
Despite the advances in VLMs' semantic grounding through self-correction, this approach trades compute for performance. Appendix~\ref{supp:tradeoff} shows the GPT-4o performance-cost tradeoff. Therefore, feedback-based reasoning becomes less practical in applications requiring low latency. Additionally, assessing VLMs' zero-shot capabilities with close-set vocabularies highlights language ambiguities. For instance, in ADE20k, similar classes like `grass', `field', `plant', and `tree' exacerbate this issue. For proprietary VLMs, we include the class list in the prompt, but this does not resolve ambiguities as each dataset may define the same classes differently. 
%For open-source VLMs, given the smaller context window, we rely on off-the-shelf embeddings for mapping, which can introduce noise.
Appendix~\ref{supp:class_mapping} provides additional quantitative analysis on the errors in class mapping.
We expect future generations of open-source VLMs to achieve significant quantitative improvements in these tasks.

\clearpage
{
    \small
    \bibliographystyle{ieeenat_fullname}
    \bibliography{main}
}

\clearpage

%\clearpage
\appendix

\section{Table of Contents}
\begin{enumerate}
    \item Sec.~\ref{supp:class_mapping} quantify the errors in class mapping.
    \item Sec.~\ref{supp:tradeoff} show the cost-performance graphs of GPT-4V and GPT-4o,
    \item Sec.~\ref{supp:prompt_tempalte} shows the prompt templates used in the experiments.
    \item Sec.~\ref{supp:dialogue} shows the high-level dialogue between the VLM agent and the Verifier.
    \item Sec.~\ref{supp:dataset_details} shows the dataset preparation and setup in our experiments.
    \item Sec.~\ref{supp:implementation_details} shows additional implementation details in our experiments.
    \item Sec.~\ref{supp:addtional_res} shows the additional experimental results. Sec~\ref{supp:feedback_accuracy} analyzes the VLM feedback in accuracy and justify that $F_1$ is a better metric to evaluate feedback quality. Sec.~\ref{supp:qualitative_res} shows the qualitative results.
\end{enumerate}

\section{Quantitative Analysis in Class Mapping Errors}
\label{supp:class_mapping}
Assessing VLMs’ zero-shot capabilities with close-set vocabularies highlights language ambiguities. In this work, we rely on off-the-shelf sentence embeddings for the class mapping. To quantify errors introduced by mapping model outputs to close-set class labels, we conducted an additional experiment: We sampled 100 raw outputs from LLaVA-1.5 in ADE. A human (one of the authors) evaluated whether the mapping from raw output to class labels, using sentence embeddings, was correct. Table~\ref{tab:error_in_mapping}
Evaluating open-vocabulary models cheaply and automatically remains an open question. Even human evaluators found 10\% of the data difficult to map correctly. We have tried to ensure fair comparisons between approaches by maintaining consistent mapping.

\begin{table}[h]
\resizebox{\columnwidth}{!}{%
\begin{tabular}{@{}ll@{}}
\toprule
Options                                                                                      & Counts \\ \midrule
The mapping is correct.                                                                      & 77     \\
The mapping is incorrect and I can provide the correct one.                                  & 13     \\
The mapping is incorrect, but it is hard to find a good one from the close set class labels. & 10     \\
Total                                                                                        & 100    \\ \bottomrule
\end{tabular}%
}
\caption{\textbf{Human studies in quantifying the error in class mapping.}}\label{tab:error_in_mapping}
\end{table}

\section{Performance-Cost Tradeoff}\label{supp:tradeoff}

\begin{figure}[h]
\includegraphics[width=\columnwidth]{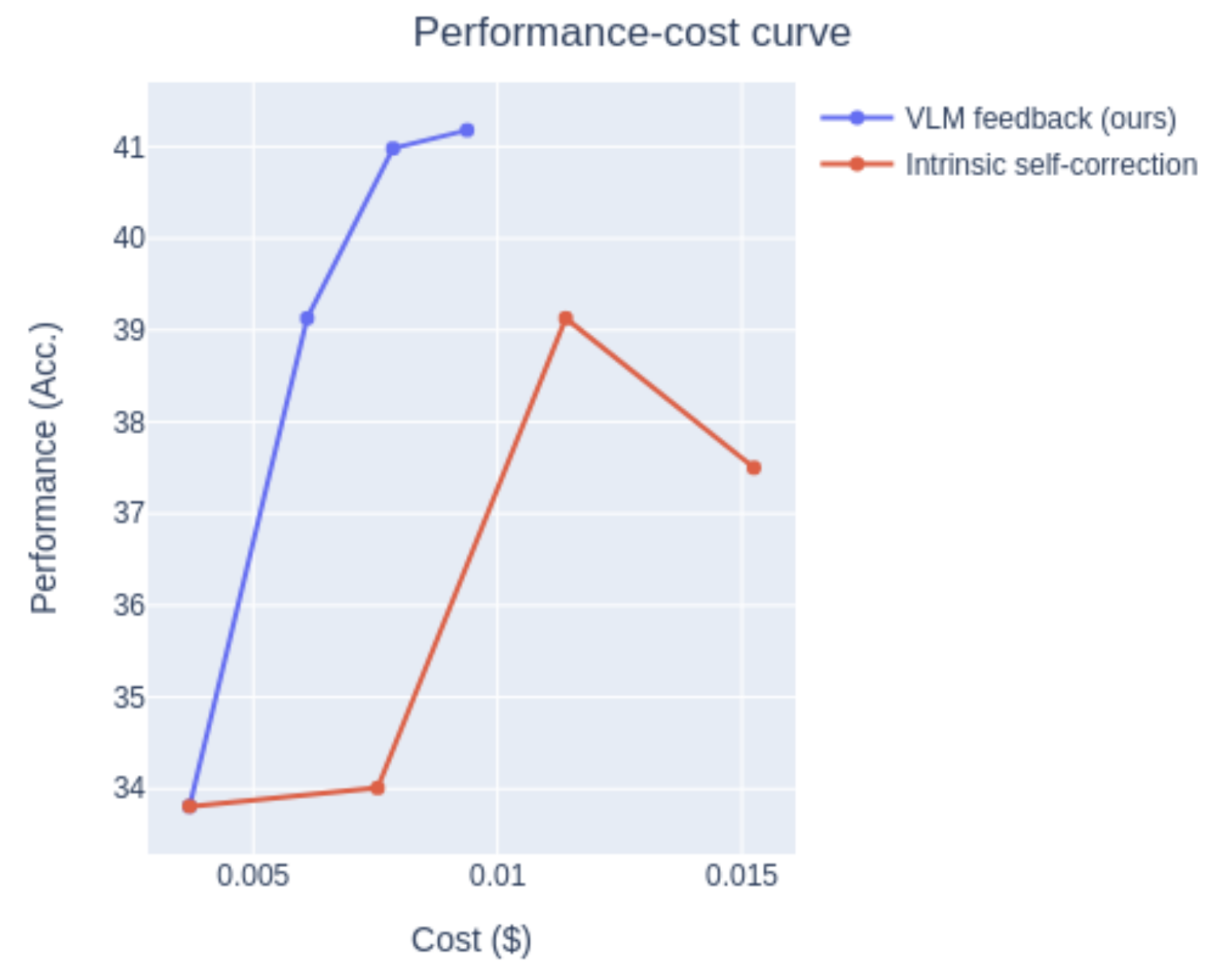}
\centering
\caption{\textbf{Cost-performance tradeoff of GPT-4o in ADE20k}}\label{fig:tradeoff}
\end{figure}

Despite the advances in VLMs' semantic grounding through self-correction, the identified self-correction trades compute for performance. Fig.~\ref{fig:tradeoff} shows the GPT-4o performance-cost tradeoff in ADE20k.

\section{Prompt Templates}\label{supp:prompt_tempalte}

We show the full prompt templates

\begin{enumerate}
    \item To producing base semantic grounding predictions in Fig.\ref{fig:prompt_template_base_predictions}
    \item To enhance previous semantic grounding predictions by taking binary feedback in Fig.~\ref{fig:prompt_template_iteartive_enhance_binary}
    \item To enhance previous semantic grounding predictions by taking class label feedback in Fig.~\ref{fig:prompt_template_iteartive_enhance_class}
    \item To produce VLM binary feedback in Fig.~\ref{fig:prompt_template_binary_vlm}.
    \item For the GPT-4V and GPT-4o experiments, we provide the class names by appending `\textit{You must answer by selecting from the following names: [COCO or ADE20k Vocabulary]}' in the prompt\footnote{\url{https://github.com/microsoft/SoM/tree/main/benchmark\#open-vocab-segmentation-on-coco}}, as shown in Fig.~\ref{fig:prompt_template_gpt4v_t0} and Fig.~\ref{fig:prompt_template_gpt4v}.

\end{enumerate}

\begin{figure*}[h]
\centering  
\begin{tcolorbox}
\begin{Verbatim}[breaklines=true, commandchars=\\\{\}]
User: You are tasked with visual semantic grounding. Your goal is to determine the class names for objects within a provided image. Each object in the image is identified by a unique ID and its location is defined by a precise bounding box, formatted as: \textbackslash{}id\{id\} \textbackslash{}box\{[x1, y1, x2, y2]\}, where coordinates specify the box corners. The inferred class name for each object is denoted as \textbackslash{}class\{class name\}. Here are the objects: \textcolor{red}{\textbackslash{}id\{2\} \textbackslash{}box\{[0.1, 0.2, 0.13, 0.43]\}}
Put your final answer by filling in the placeholder(s) in the following string at the beginning: "\textcolor{red}{\textbackslash{}id\{2\} \textbackslash{}box\{[0.1, 0.2, 0.13, 0.43]\} \textbackslash{}class\{your answer here\}}"
\end{Verbatim}
\end{tcolorbox}
\caption{Prompt template to produce the base predictions. The text in red represents variables.}\label{fig:prompt_template_base_predictions}
\end{figure*}

\begin{figure*}[h]
\centering  
\begin{tcolorbox}
\begin{Verbatim}[breaklines=true, commandchars=\\\{\}]
User: You are tasked with visual semantic grounding. Your goal is to determine the class names for objects within a provided image and leverage the insights from expert analyses. The expert analyses offer detailed information on the inferred class names for each object in the provided image. Each object in the image is identified by a unique ID and its location is defined by a precise bounding box, formatted as: \textbackslash{}id\{id\} \textbackslash{}box\{[x1, y1, x2, y2]\}, where coordinates specify the box corners. The inferred class name for each object is denoted as \textbackslash{}class\{class name\}. I have labeled each object with its ID and overlaid its segmentation mask on the image to clarify the correspondences.

One expert analyses on the provided image are shown below:
* Analysis 1
Object(s) with inferred class names: \textcolor{red}{\textbackslash{}id\{2\} \textbackslash{}box\{[0.1, 0.2, 0.13, 0.43]\} \textbackslash{}class\{wall\}}
Expert's decision(s) on class names: The inferred class name(s) for \{incorrect obj id\} are incorrect. The inferred class name(s) for \textcolor{red}{\textbackslash{}id\{2\}} are not "\textcolor{red}{wall}". 
Expert's suggestion: Adjust the class names for objects with IDs \textcolor{red}{\textbackslash{}id\{2\}}

Examine the image and the expert analyses to determine the true class name of the object(s): \textcolor{red}{\textbackslash{}id\{2\} \textbackslash{}box\{[0.1, 0.2, 0.13, 0.43]\}}. Put your final answer by filling in the placeholder(s) in the following string at the beginning: "\textcolor{red}{\textbackslash{}id\{2\} \textbackslash{}box\{[0.1, 0.2, 0.13, 0.43]\} \textbackslash{}class\{your answer here\}}"
\end{Verbatim}
\end{tcolorbox}
\caption{Prompt template to improve semantic grounding predictions by taking Binary Feedback. The text in red represents variables.}\label{fig:prompt_template_iteartive_enhance_binary}
\end{figure*}

\begin{figure*}[h]
\centering  
\begin{tcolorbox}
\begin{Verbatim}[breaklines=true, commandchars=\\\{\}]
User: You are tasked with visual semantic grounding. Your goal is to determine the class names for objects within a provided image and leverage the insights from expert analyses. The expert analyses offer detailed information on the inferred class names for each object in the provided image. Each object in the image is identified by a unique ID and its location is defined by a precise bounding box, formatted as: \textbackslash{}id\{id\} \textbackslash{}box\{[x1, y1, x2, y2]\}, where coordinates specify the box corners. The inferred class name for each object is denoted as \textbackslash{}class\{class name\}. I have labeled each object with its ID and overlaid its segmentation mask on the image to clarify the correspondences.

One expert analyses on the provided image are shown below:
* Analysis 1
Object(s) with inferred class names: \textcolor{red}{\textbackslash{}id\{2\} \textbackslash{}box\{[0.1, 0.2, 0.13, 0.43]\} \textbackslash{}class\{wall\}}
Expert's decision(s) on class names: The inferred class name(s) for \textcolor{red}{\textbackslash{}id\{2\}} are incorrect. The inferred class name(s) for \textcolor{red}{\textbackslash{}id\{2\}} are not "\textcolor{red}{wall}". 
Expert's suggestion: Adjust the class names for objects with IDs \textcolor{red}{\textbackslash{}id\{2\}} to \textcolor{red}{\textbackslash{}class\{rail\}}.

Examine the image and the expert analyses to determine the true class name of the object(s): \textcolor{red}{\textbackslash{}id\{2\} \textbackslash{}box\{[0.1, 0.2, 0.13, 0.43]\}}. Put your final answer by filling in the placeholder(s) in the following string at the beginning: "\textcolor{red}{\textbackslash{}id\{2\} \textbackslash{}box\{[0.1, 0.2, 0.13, 0.43]\} \textbackslash{}class\{your answer here\}}"
\end{Verbatim}
\end{tcolorbox}
\caption{Prompt template to improve semantic grounding predictions by taking Class Label Feedback. The text in red represents variables.}\label{fig:prompt_template_iteartive_enhance_class}
\end{figure*}

\begin{figure*}[h]
\centering  
\begin{tcolorbox}
\begin{Verbatim}[breaklines=true, commandchars=\\\{\}]
User: Does this \textcolor{red}{cropped image} contain "\textcolor{red}{wall}"? Answer yes or no.
\end{Verbatim}
\end{tcolorbox}
\caption{Prompt template to derive VLM binary feedback. The text in red represents variables.}\label{fig:prompt_template_binary_vlm}
\end{figure*}

\begin{figure*}[h]
\centering  
\begin{tcolorbox}
\begin{Verbatim}[breaklines=true, commandchars=\\\{\}]
User: I have labeled a bright numeric ID at the center for each visual object in the image. Please enumerate their names. You must answer by selecting from the following names: \textcolor{red}{[Class list]}
\end{Verbatim}
\end{tcolorbox}
\caption{Prompt template for GPT-4V and GPT-4o to produce the base predictions. Following prior work~\citep{som_prompt}, we include the full class list in the text prompt. The text in red represents variables.}\label{fig:prompt_template_gpt4v_t0}
\end{figure*}

\begin{figure*}[h]
\centering  
\begin{tcolorbox}
\begin{Verbatim}[breaklines=true, commandchars=\\\{\}]
User: You are tasked with visual semantic grounding. Your goal is to determine the class names for objects within a provided image and leverage the insights from expert analyses. The expert analyses offer detailed information on the inferred class names for each object in the provided image. Each object in the image is identified by a unique ID and its location is defined by a precise bounding box, formatted as: \textbackslash{}id\{id\} \textbackslash{}box\{[x1, y1, x2, y2]\}, where coordinates specify the box corners. The inferred class name for each object is denoted as  \textbackslash{}class\{class name\}. I have labeled each object with its ID and overlaid its segmentation mask on the image to clarify the correspondences.

One expert analyses on the provided image are shown below:
* Analysis 1
Object(s) with inferred class names: \textcolor{red}{\textbackslash{}id\{2\} \textbackslash{}box\{[0.1, 0.2, 0.13, 0.43]\} \textbackslash{}class\{wall\}}
Expert's decision(s) on class names: The inferred class name(s) for \{incorrect obj id\} are incorrect. The inferred class name(s) for \textcolor{red}{\textbackslash{}id\{2\}} are not "\textcolor{red}{wall}". 
Expert's suggestion: Adjust the class names for objects with IDs \textcolor{red}{\textbackslash{}id\{2\}}

Examine the image and the expert analyses to determine the true class name of the object(s): \textcolor{red}{\textbackslash{}id\{2\} \textbackslash{}box\{[0.1, 0.2, 0.13, 0.43]\}}. Put your final answer by filling in the placeholder(s) in the following string at the beginning: "\textcolor{red}{\textbackslash{}id\{2\} \textbackslash{}box\{[0.1, 0.2, 0.13, 0.43]\} \textbackslash{}class\{your answer here\}}"

You must answer by selecting from the following names: \textcolor{red}{[ADE Class List]}
\end{Verbatim}
\end{tcolorbox}
\caption{Prompt template for GPT-4V to improve semantic grounding predictions by taking Binary Feedback. Following prior work~\cite{som_prompt}, we include the full class list in the text prompt. The text in red represents variables.}\label{fig:prompt_template_gpt4v}
\end{figure*}

\section{Example Dialogue}\label{supp:dialogue}
\begin{figure*}[t]
\includegraphics[width=\textwidth]{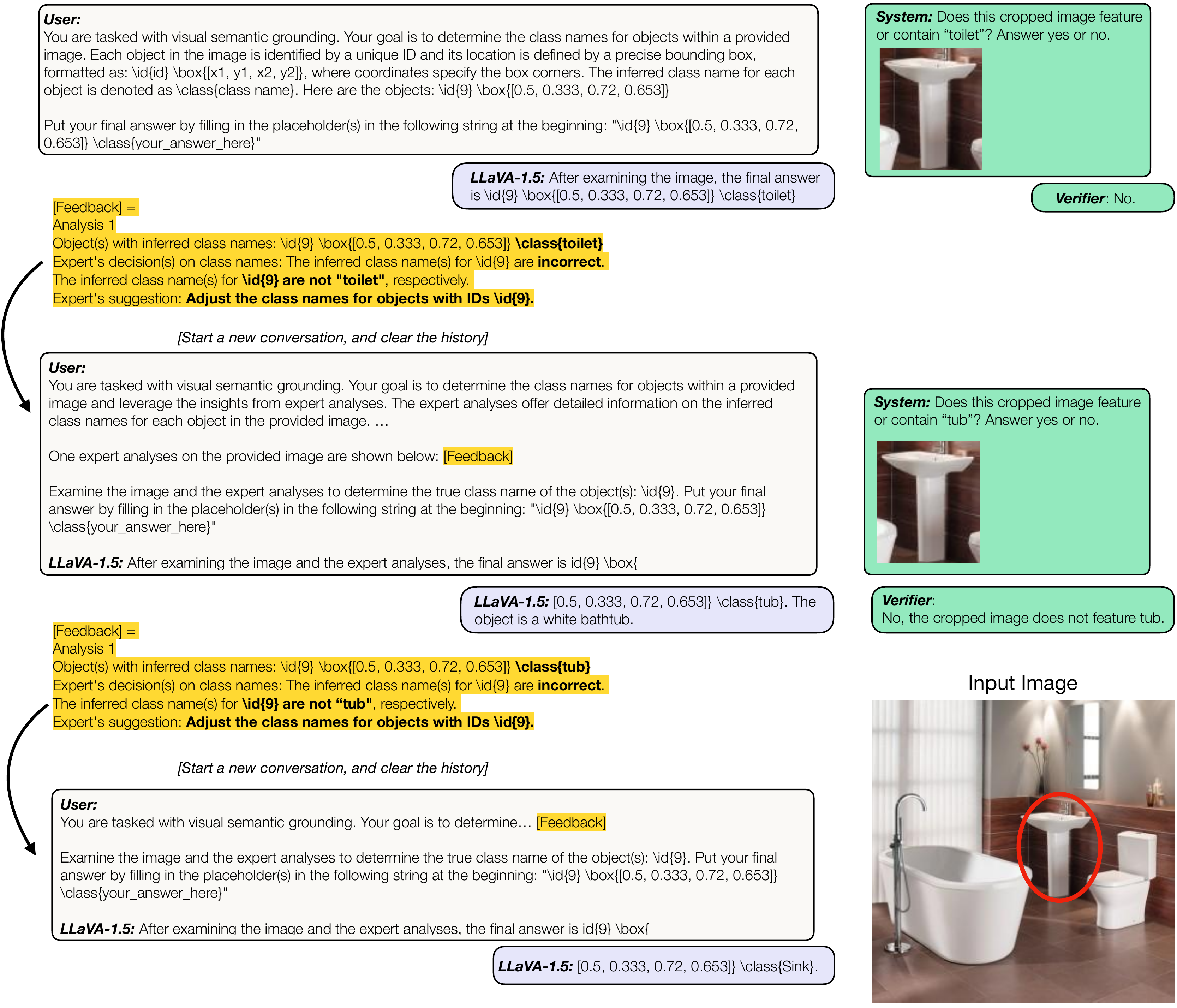}
\caption{\textbf{Example dialogue of using iterative binary self-feedback to improve semantic grounding in VLMs.} 
Initially, the user queries the semantic class of an object within a bounding box. At the first attempt, the VLM responds without feedback. To refine accuracy, we use the verifier (the same VLM) to answer a yes/no question for binary self-feedback. Incorporating this feedback, we prompt the VLM again, leading to a refined prediction. The VLM's initial guess evolves from `toilet' to `bathtub', and ultimately to `sink' -- the correct classification.
%From top to bottom, the user ask for the semantic class of the object within the bounding box. For the first timestep, the VLM answers without any additional feedback. We prompt the verifier (using the exact same VLM) an yes/no question to obtain the binary self-feedback. Then, we prompt the VLM again with the binary self-feedback to produce a more accurate prediction. Overall, the VLM prediction changes from toilet to bathtub, and finally to sink, which is the correct answer.
}\label{fig:qualitative_res}
\end{figure*}

\begin{figure*}[t]
\includegraphics[width=\textwidth]{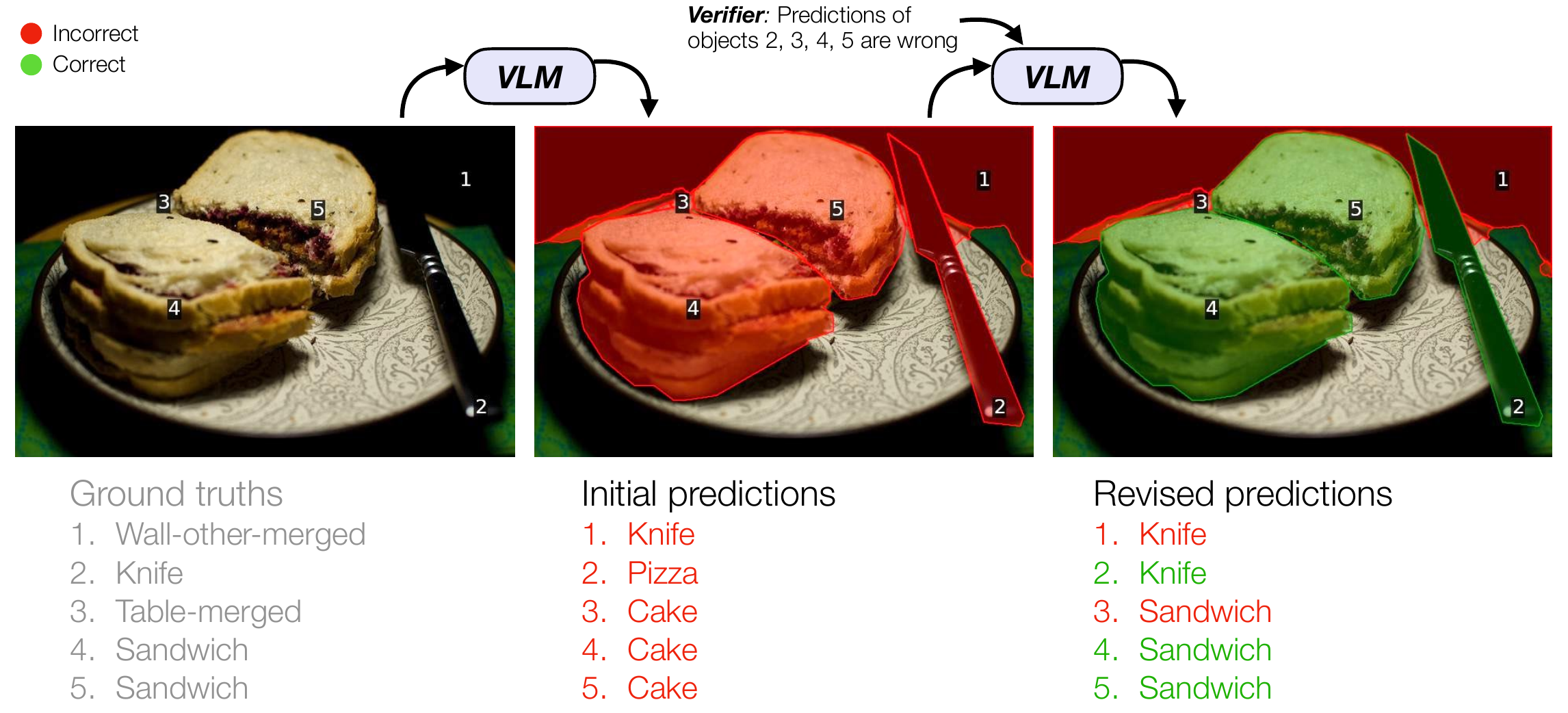}
\caption{\textbf{Enhancing semantic grounding in VLMs with self-generated feedback.} We use GPT-4V as the VLM here. From the left to the center figure, GPT-4V takes the SoM-prompted image~\cite{som_prompt} as input and struggles to predicts the class names of each object. From the center to the right figure, GPT-4V takes the same SoM-prompted image and the additional feedback from the verifier and successfully correct the class names of three out of five objects. The verifier is another GPT-4V that operates on an altered input image 
% \& SoM and 
and may produce noisy feedback, \eg, misclassify object 1 as correct.
%~\footnotemark
}
\label{fig:gpt4v_qualitative}
\end{figure*}

In Fig.~\ref{fig:qualitative_res}, we demonstrate the iterative interactions between a VLM agent and the Verifier. In Fig.~\ref{fig:gpt4v_qualitative}, we show the effectiveness of VLM binary verification in GPT-4V~\footnote{GPT-4V predictions with simplified prompts as of Mar 22, 2024: \url{https://imgur.com/a/nbKjIlb}}.

\section{Dataset Details}\label{supp:dataset_details}
We use ADE20k and COCO panoptic segmentation dataset to evaluate the semantic grounding performance in VLMs. We adopt SoM split provided in the prior work~\cite{som_prompt}\footnote{\url{https://github.com/microsoft/SoM/tree/main/benchmark\#dataset}}.
ADE20k is a large-scale dataset with fine-grained segmentation labels. We adopt the variant with 150 classes, commonly referred to as ADE20k-150. COCO panoptic segmentation is a standard dataset to evaluate visual grounding. There are 133 fine-grained classes in total, composed of 80 thing classes and 53 stuff classes.
Consistent with prior works, SoM~\citep{som_prompt}, we use the same subset of 100 images for ADE20k and COCO for our analysis.
There are 100 images and 488 segmentation masks in ADE20k SoM split and 101 and 628 segmentation masks in COCO SoM split.

Every region $\rr_\ri$ in ADE20k and COCO panoptic segmentation dataset is represented with segmentation mask. We convert them to a more compact representation, \ie bounding box, and feed them to the VLMs in the text prompt

\section{Implementation Details}\label{supp:implementation_details}

Every experiment throughout this paper is run over three seeds and we report the average scores except for experiments with proprietary VLMs. All the experiments are run in a single-node machine with two A40 GPUs.
\new{In the experiments with binary or class label feedback, we only ask VLMs to correct those that are incorrect based on the feedback. Therefore, if the feedback is noisy, \eg VLM binary verification, VLMs can possibly decrease the performances. See Fig.~\ref{fig:llava_qualitative_res_2_fail} for example.}

%Every region $\rr_\ri$ in ADE20k and COCO panoptic segmentation dataset is represented with segmentation mask. We convert them to a more compact representation, \ie bounding box, and feed them to the VLMs in the text prompt. We follow ~\cite{som_prompt} to set the alpha parameters in SoM as 0.2 and 0.4 in ADE20k and COCO, respectively.

\textbf{Open-source VLMs.} We adopt LLaVA-1.5 13b (from \url{https://huggingface.co/llava-hf/llava-1.5-13b-hf}), ViP-LLaVA 13b (from \url{https://huggingface.co/llava-hf/vip-llava-13b-hf}), and CogVLM (from \url{https://huggingface.co/THUDM/CogVLM}).
When perform the VLM forward pass $\ro_\ri = \text{VLM}(\rx, \rr_\ri, \rq)$, we set the temperature to 0.9, top\_p to 0.8, max\_new\_tokens to 1024, and draw five samples per forward pass. We take the majority vote responses as the final answers $\ro_\ri$.

\textbf{GPT-4V.} As suggested in prior work~\citep{som_prompt, gpt4v_report}, GPT-4V exhibits better grounding ability when the objects are specified by visual prompts rather than text prompts. Therefore, we adopt GPT-4V \& SoM to obtain the base predictions, where we overlay object masks and numeric identifiers on the images. Furthermore, when using VLMs to produce feedback, we apply SoM to specify each object. 
Finally, since GPT-4V has a longer context window compared to open-source VLMs, we include the class list in the prompt to encourage better alignment between the responses and the ground truth. All GPT-4V experiments are done over the OpenAI API and we follow the exact same evaluation procedures%described in Sec.~\ref{sec:exp_protocols}
, where we use the off-the-shelf text embeddings~\citep{all-mpnet-base-v2} to map the GPT-4V outputs $\ro_\ri$ to the nearest label from the class label list. 

We follow the implementation provided in \cite{som_prompt}\footnote{\url{https://github.com/microsoft/SoM/blob/main/gpt4v.py}} and set the system prompt as: \textit{- For any marks mentioned in your answer, please highlight them with [].} We follow~\cite{som_prompt} to set the alpha parameters in SoM as 0.2 and 0.4 in ADE20k and COCO, respectively. We use the endpoint \texttt{gpt-4-0125-preview}.
%We assessed the GPT-4V model in March 2024 via OpenAI API.

\textbf{GPT-4o.} Similar to GPT-4V, we empirically found that SoM prompts improve the base predictions in the semantic grounding tasks in ADE20k. We, therefore, hypothesize that GPT-4o benefits by having SoM prompts. We use the endpoint \texttt{gpt-4o-2024-05-13}.
%We assessed the GPT-4o model in May 2024 via OpenAI API.

\section{Additional Results}\label{supp:addtional_res}

\subsection{Feedback Accuracy does not Strongly Correlate with Semantic Grounding with Iteratively Self-Generated Feedback}~\label{supp:feedback_accuracy}

In the main paper, we measure feedback in $F_1$ score. Another intuitive evaluation metric is feedback accuracy, denoted as $Acc_\text{feedback}$ and we show the results in Table~\ref{tab:provide_feedback_acc}.
We find that VLM binary verification with a higher $Acc_\text{feedback}$ \emph{does not necessary} lead to a higher grounding accuracy in the iterative setup. % in Sec.~\ref{sec:iterative_feedback}. 
On average, we find that $Acc_\text{feedback}$ achieve an 0.11 Spearman rank correlation coefficient~\cite{spearman} with grounding accuracy at $\rt=3$ as compared to 0.72 achieved by $F_1$. We conclude that $F_1$ is a better evaluation metric for measure feedback quality in this work.

%Acc table
\begin{table}[t]
\centering
\footnotesize
\resizebox{\columnwidth}{!}{%
\begin{tabular}{@{}lrrrrr@{}}
\toprule
%\multicolumn{1}{c}{} & \multicolumn{1}{c}{Visual prompt} &  \multicolumn{1}{c}{Zero-shot self-consistency} & \multicolumn{1}{c}{LLaVA-1.5} & \multicolumn{1}{c}{Vip-LLaVA}  & \multicolumn{1}{c}{CogAgent} \\ \midrule
\multicolumn{1}{l}{} & \multicolumn{1}{r}{Visual prompt}  & \multicolumn{1}{c}{LLaVA-1.5} & \multicolumn{1}{c}{ViP-LLaVA}  & \multicolumn{1}{c}{CogVLM} \\ \midrule
\rowcolor[HTML]{EFEFEF} Intrinsic Self-Correction       & N/A    & 47.03                        & 47.13                               &                              \textbf{59.5}\\ 
\cmidrule(l){2-5} 
\multirow{3}{*}{VLM Binary Verification}& Visual marks           & 55.5                        &                       65.2&                              52.3\\
                                        & RoI crop               & \textbf{64.1}               &                                57.6&                     57\\
                                        & Visual marks + RoI crop&                         62.1&                                 \textbf{67.2}&                              52.9\\  \bottomrule
\end{tabular}%
}
\caption{\textbf{Accuracy of the VLMs binary feedback $Acc_\text{feedback}$.} We find that intrinsic self-correction often improves accuracy in VLMs with lower base prediction performance due to imbalanced oracle binary feedback.}
\label{tab:provide_feedback_acc}
\vspace{-3mm}
\end{table}

\subsection{Qualitative Results}\label{supp:qualitative_res}

\begin{figure*}[t]
\includegraphics[width=\textwidth]{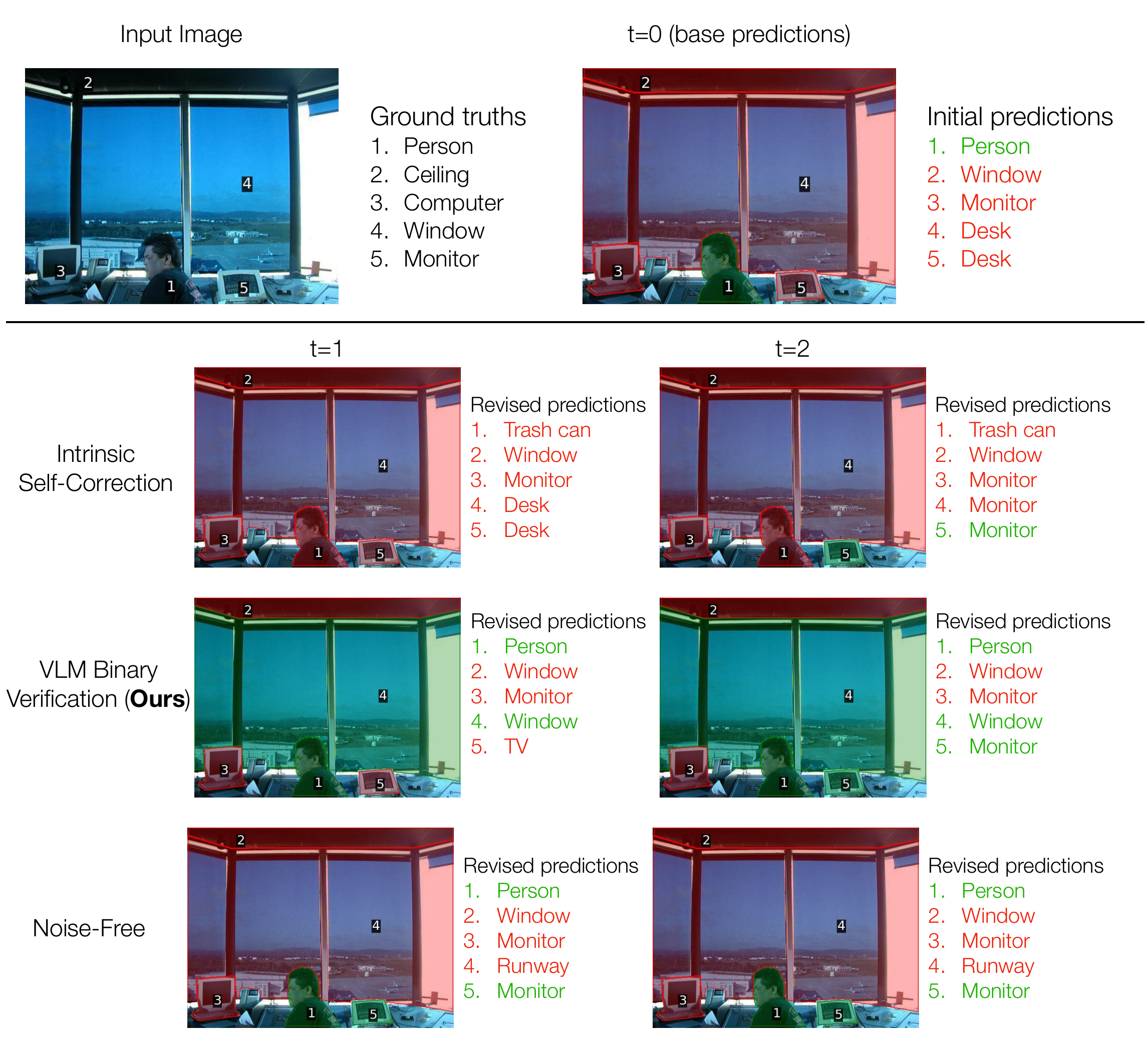}
\caption{\textbf{LLaVA-1.5 qualitative results in ADE20k.} We visualize the predictions of LLaVA-1.5 at time steps from 0 to 2. Intrinsic self-correction fails to identify which predictions are correct/incorrect, while VLM binary verification and Noise-free feedback provide explicit signal on each region, leading to a better chance of correction. 
%Note that we draw multiple samples in the VLM forward pass, therefore, leading to slightly different results even when the image and query are the same (See Appendix~\ref{supp:implementation_details}).
From $\rt=0$ to $\rt=1$, we find that VLM might produce different results (object 4) even when receiving the same feedback (VLM binary verification and Noise-free). As explained in Appendix~\ref{supp:implementation_details}, in the VLMs forward pass, we draw multiple sequences and take the majority vote as the final responses.
For the sake of visualization, we put a bright ID on each object and highlight the incorrect predictions in red and the correct predictions in green.}\label{fig:llava_qualitative_res_1}
\end{figure*}

\begin{figure*}[t]
\includegraphics[width=\textwidth]{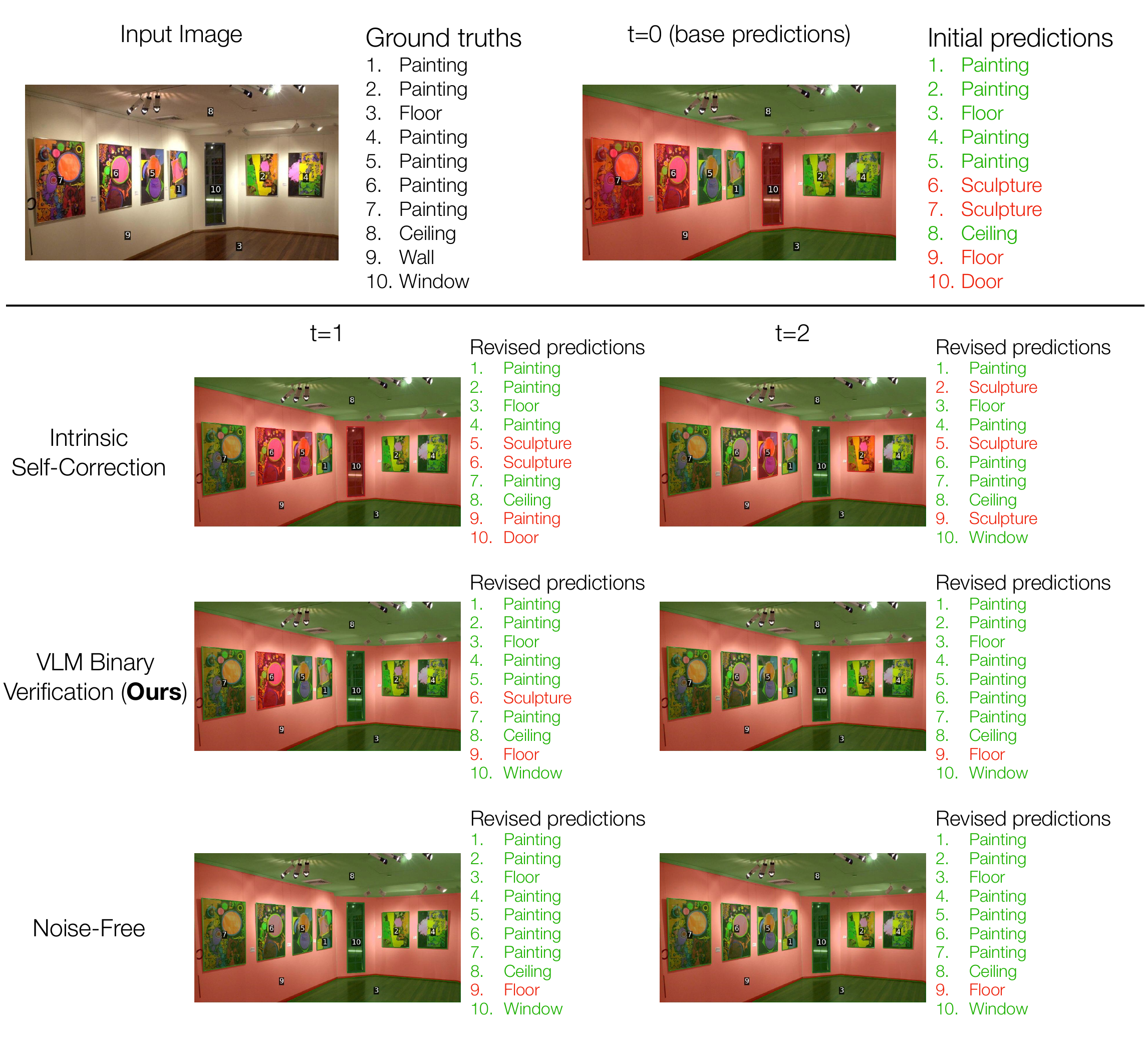}
\caption{\textbf{ViP-LLaVA qualitative results in ADE20k.} We visualize the predictions of ViP-LLaVA at time steps from 0 to 2. Intrinsic self-correction fails to identify which predictions are correct/incorrect, while VLM binary verification and Noise-free  feedback provide explicit signal on each region, leading to a better chance of correction. Note that we draw multiple samples in the VLM forward pass, therefore, leading to slightly different results even when the image and query are the same (See Appendix~\ref{supp:implementation_details}).
For the sake of visualization, we put a bright ID on each object and highlight the incorrect predictions in red and the correct predictions in green.}\label{fig:vip_llava_qualitative_res_1}
\end{figure*}

\begin{figure*}[t]
\includegraphics[width=\textwidth]{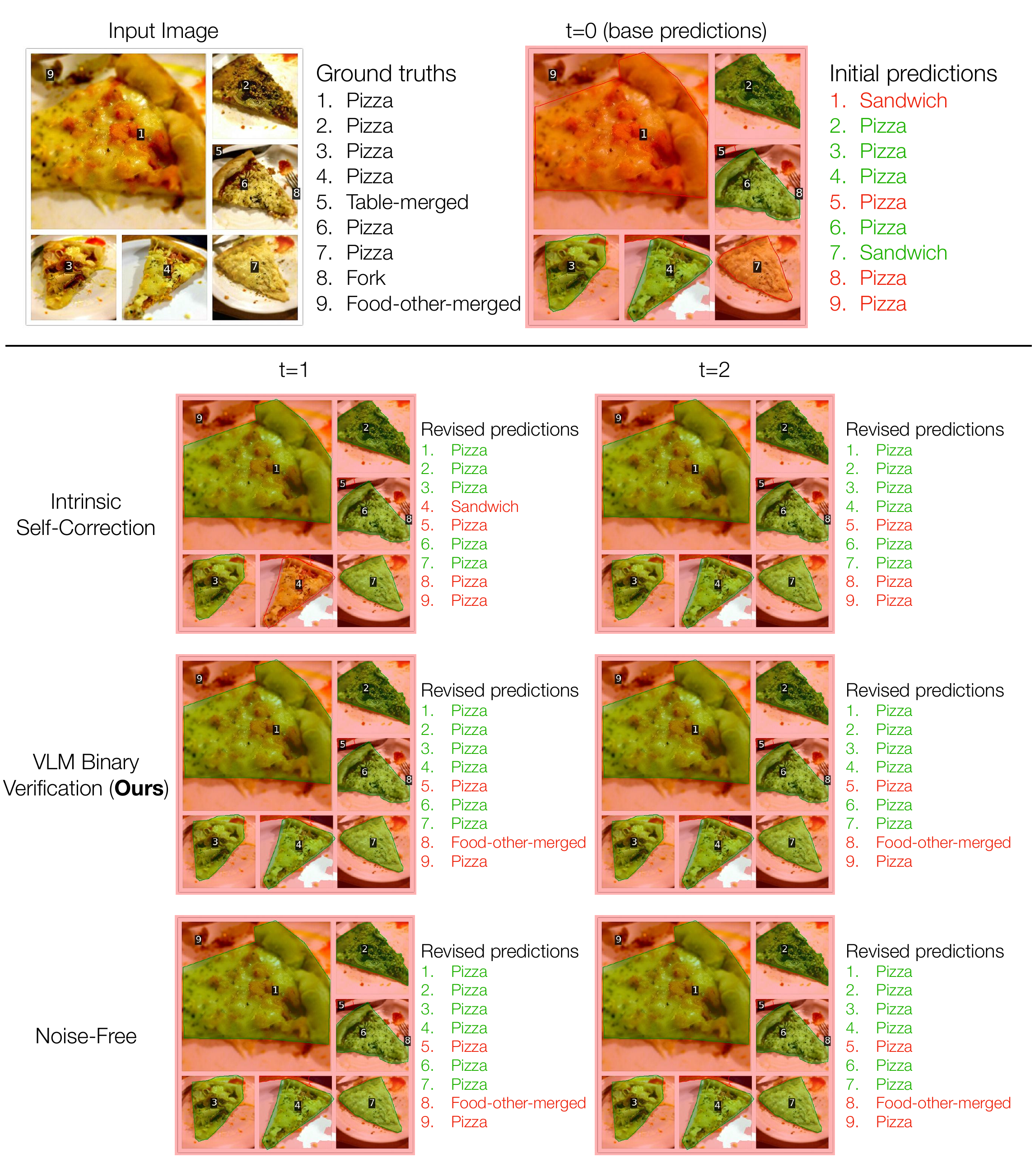}
\caption{\textbf{CogVLM qualitative results in COCO.} We visualize the predictions of CogVLM at time steps from 0 to 2. 
For the sake of visualization, we put a bright ID on each object and highlight the incorrect predictions in red and the correct predictions in green.}\label{fig:cogvlm_qualitative_res_1}
\end{figure*}

\begin{figure*}[t]
\includegraphics[width=\textwidth]{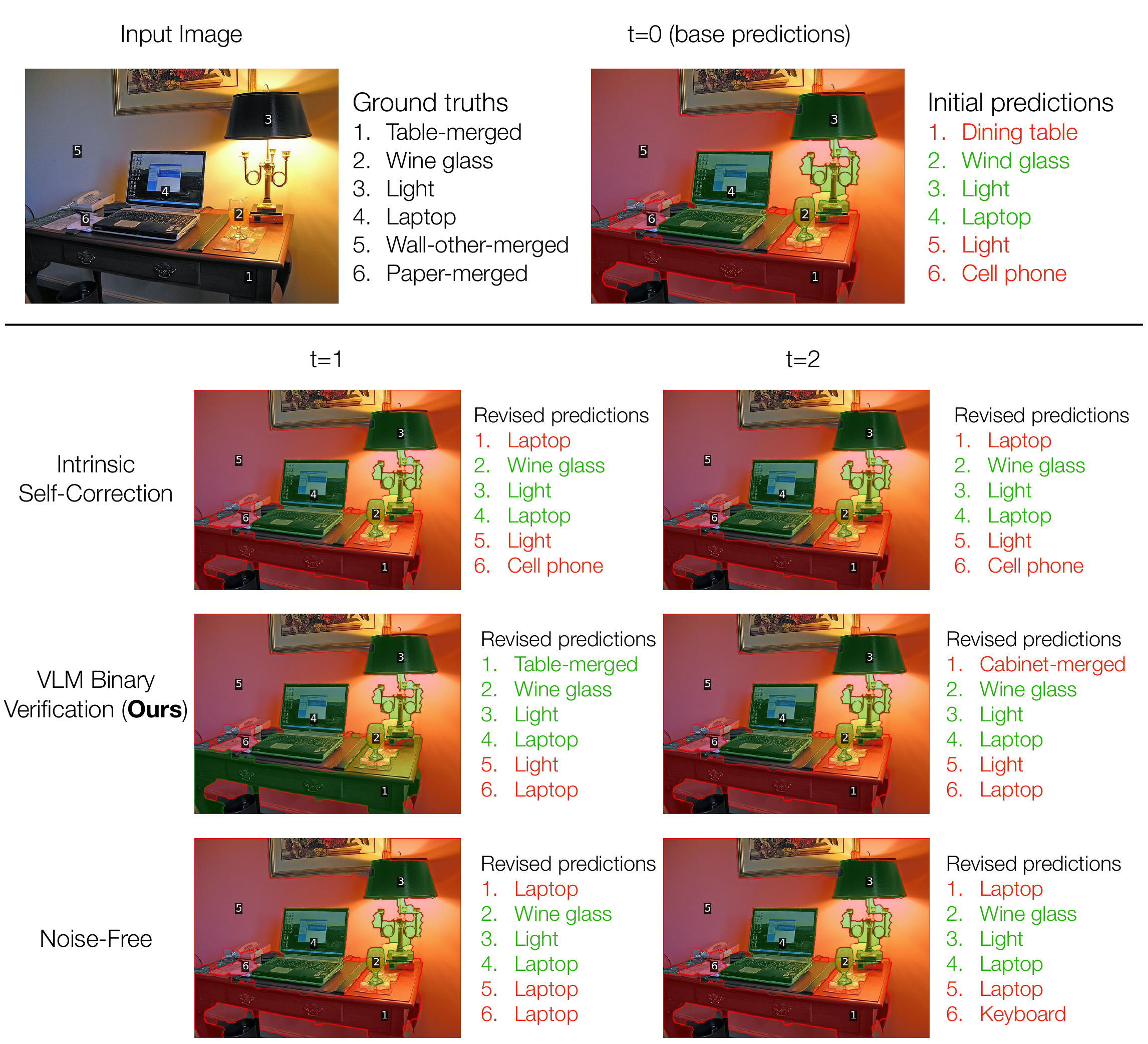}
\caption{\textbf{[Failure case study] LLaVA-1.5 qualitative results in COCO.} All three approaches cannot fix the errors in the initial predictions. 
For VLM binary verification, from $\rt=1$ to $\rt=2$, the predictions changes from correct (table-merged) to incorrect (cabinet-merged) since the VLM verifier is not perfect and, therefore, providing misleading feedback.
Even with the noise-free feedback, LLaVA-1.5 struggle to adjust the predictions.
For the sake of visualization, we put a bright ID on each object and highlight the incorrect predictions in red and the correct predictions in green.}\label{fig:llava_qualitative_res_2_fail}
\end{figure*}

We share additional qualitative results on ADE20k and COCO in Fig.~\ref{fig:llava_qualitative_res_1}, Fig.~\ref{fig:vip_llava_qualitative_res_1}, Fig.~\ref{fig:cogvlm_qualitative_res_1}. We also note that most of the failure cases occur when 1) the VLMs keep their own predictions even though the feedback refers them as incorrect predictions or 2) when the self-generated feedback is incorrect, as shown in Fig.~\ref{fig:llava_qualitative_res_2_fail}.

%Fig.~\ref{fig:llava_qualitative_res_1} and Fig.~\ref{fig:llava_qualitative_res_2} showcase the results of LLaVA and we find that intrinsic self-correction struggles to adjust the predictions, which reflects on the low $F_1$ scores in Table~\ref{tab:provide_feedback_f1}. Fig.~\ref{fig:gpt4v_qualitative_res_1} showcase the results of GPT-4V \& SoM and we observe that GPT-4V follows the feedback better than other open-source VLMs. Fig.~\ref{fig:vip_llava_qualitative_res_1} and Fig.~\ref{fig:vip_llava_qualitative_res_2} showcase the results of ViP-LLaVA and we find that in some examples, VLM binary verification produces better reuslts than noise-free feedback due to the randomness mentioned in Appendix~\ref{supp:implementation_details}. Fig.~\ref{fig:cogvlm_qualitative_res_1} and Fig.~\ref{fig:cogvlm_qualitative_res_2} showcase the results of CogVLM and we find that both intrinsic self-correction and VLM binary verification struggle to adjust the predictions, potentially due to the poor feedback performance shown in Table~\ref{tab:provide_feedback_f1}.  

\end{document}